\documentclass[journal]{IEEEtran}
\usepackage{times}
\usepackage{epsfig}
\usepackage{graphicx}
\usepackage{amsmath}
\usepackage{amssymb}
\usepackage{xspace}

\usepackage{subfigure}
\usepackage{booktabs}
\usepackage{multirow}

\usepackage{algorithm,algpseudocode}
\usepackage{color}
\usepackage{graphics}

\usepackage{array}
\usepackage{comment}

\usepackage[colorlinks,linkcolor=blue]{hyperref}

\newcommand{\etal}{\textit{et al}.\xspace}
\newcommand{\etc}{\textit{etc}\xspace}
\newcommand{\ie}{\textit{i}.\textit{e}.\xspace}
\newcommand{\eg}{\textit{e}.\textit{g}.\xspace}

\newcommand{\tabincell}[2]{\begin{tabular}{@{}#1@{}}#2\end{tabular}}

\def\qcr{\fontfamily{qcr}\selectfont}

\def\qi{\textcolor{black}}
\def\qii{\textcolor{black}}

\def\myblue{\textcolor{black}}

\ifCLASSINFOpdf
\else
\fi
\begin{document}
	
	%
	\title{Attention-guided Context Feature Pyramid Network for Object Detection}
	%
	%
	%
	
	
	\author{
		Junxu Cao$^*$\thanks{$^*$Authors contributed equally.},
		Qi Chen$^*$,
		Jun Guo,
		and Ruichao Shi
		\\
		\thanks{Junxu Cao, Jun Guo, and Ruichao Shi are with the Tencent. E-mail: \{qibaicao, garryshi\}@tencent.com, artanis.protoss@outlook.com}
		\thanks{Qi Chen is with the School of Software Engineering, South China University of Technology. E-mail: sechenqi@mail.scut.edu.cn}
	}

	\maketitle
	
	\begin{abstract}
		For object detection, how to address the contradictory requirement between feature map resolution and receptive field on high-resolution inputs still remains an open question. In this paper, to tackle this issue, we build a novel architecture, called Attention-guided Context Feature Pyramid Network (AC-FPN), that exploits discriminative information from various large receptive fields via integrating attention-guided multi-path features. The model contains two modules. The first one is Context Extraction Module (CEM) that explores large contextual information from multiple receptive fields. As redundant contextual relations may mislead localization and recognition, we also design the second module named Attention-guided Module (AM), which can adaptively capture the salient dependencies over objects by using the attention mechanism. AM consists of two sub-modules, i.e., Context Attention Module (CxAM) and Content Attention Module (CnAM), which focus on capturing discriminative semantics and locating precise positions, respectively. Most importantly, our AC-FPN can be readily plugged into existing FPN-based models. Extensive experiments on object detection and instance segmentation show that existing models with our proposed CEM and AM significantly surpass their counterparts without them, and our model successfully obtains state-of-the-art results. We have released the source code at: \href{https://github.com/Caojunxu/AC-FPN}{https://github.com/Caojunxu/AC-FPN}.
	\end{abstract}
	

	%
	\IEEEpeerreviewmaketitle
	
	
	\section{Introduction}
	%
	%
	%
	%

\begin{figure}[t]
		\begin{center}
			\includegraphics[width=1.0\linewidth]{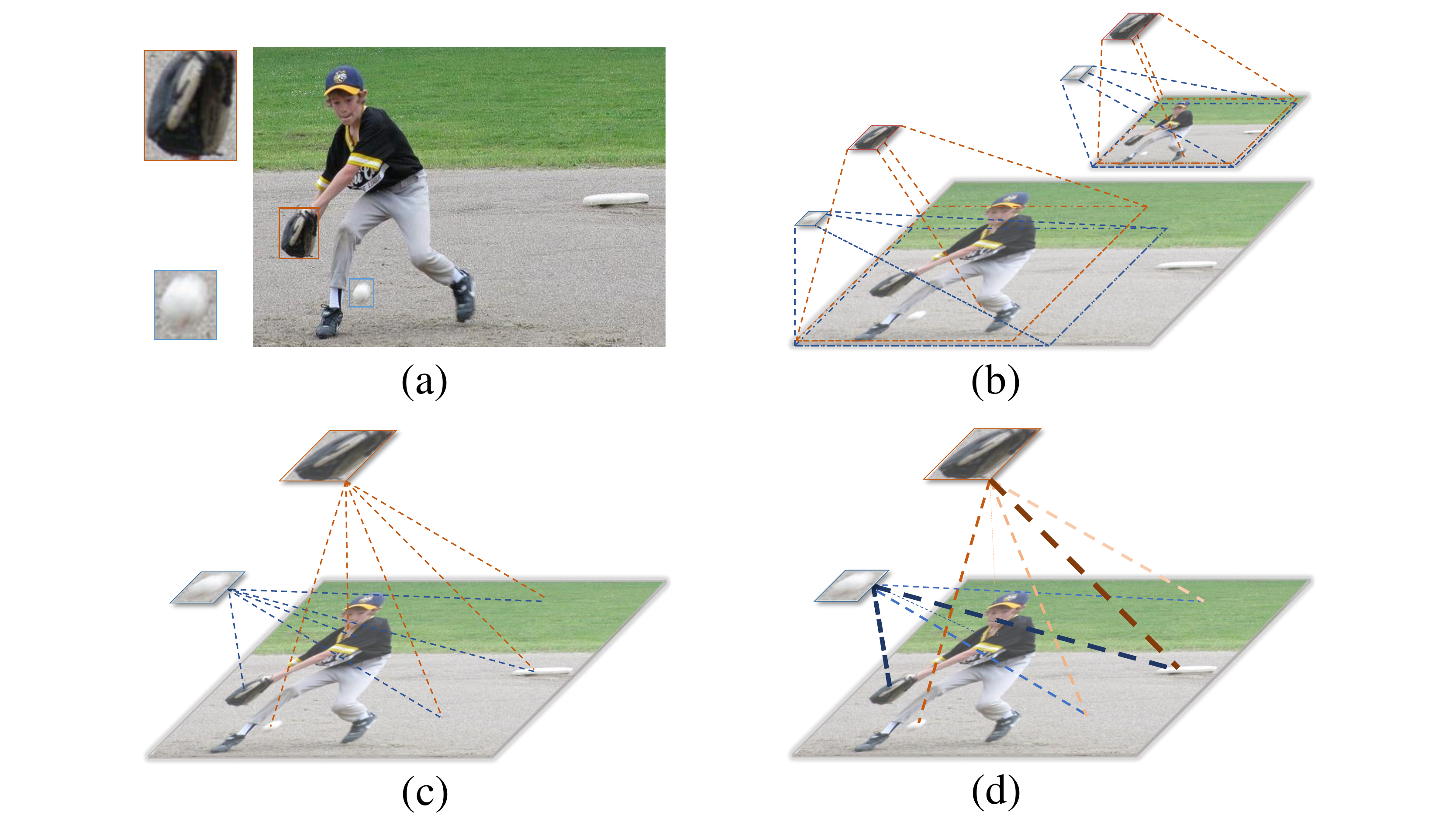}
		\end{center}
		\caption{
			(a) Detected objects.
			(b) The receptive fields of the same model on images of different sizes.
			(c) Captured context information from various receptive fields. 
			(d) Identified salient relations.
			The dashed lines indicate the context dependencies over images and the line weight refers to the degree of correlation.
		}
		\label{fig:samples}
	\end{figure}

	\IEEEPARstart{O}{bject} detection is a fundamental but non-trivial problem in computer vision (Fig.~\ref{fig:samples}(a)).
	The study of this task can be applied to \myblue{various} applications, such as face detection~\cite{jiang2017face,yang2016wider,yang2018faceness}, people counting~\cite{shami2018people,stewart2016end}, pedestrian detection~\cite{wang2018pedestrian,li2017scale,qian2019oriented}, object tracking~\cite{feichtenhofer2017detect,yuan2017incremental,chen2018learning,hu2018robust,wang2018learning}, \etc. However, how to \myblue{perform} the task effectively still remains an open question.

	Nowadays, to accurately locate objects, representative object detectors, \eg, Faster R-CNN~\cite{ren2015faster}, RetinaNet~\cite{lin2017focal}, and DetNet~\cite{li2018detnet}, use high-resolution images (with the shorter edge being $800$) as inputs, which contain much more detailed information 
	\qii{and improve the performance in object detection (See AP in Table~\ref{tab:impact_image_size}).}
	However, unfortunately, images with higher resolution require neurons to have larger receptive fields to obtain \myblue{effective} semantics (Fig.~\ref{fig:samples}(b)).
	\qii{Otherwise, it will deteriorate the performance when capturing large objects in the higher resolution images. (See $\text{AP}_L$ in Table~\ref{tab:impact_image_size}).}

	Intuitively, to obtain a larger receptive field, we can design a deeper network model by increasing the convolutional and downsampling layers, \myblue{where the downsampling layer includes the pooling layer and the convolutional layer with a stride larger than $1$.}
	However, simply increasing the number of convolutional layers is rather inefficient. It leads to much more parameters and thereby causes higher computational and memory costs. What's worse, aggressively deep networks are hard to optimize due to the overfitting problem~\cite{he2016deep}. On the other hand, the increased number of downsampling layers results in reduced feature map sizes, which causes more challenging issues in localization. Thus, how to \myblue{build a model} that can achieve large receptive fields while maintaining high-resolution feature maps remains a key issue in object detection.
	
	\myblue{Recently}, FPN~\cite{lin2017feature} is proposed to exploit the inherent multi-scale feature representation of deep convolutional networks. 
	More specifically, by introducing a top-down pathway, FPN combines low-resolution, large-receptive-field features with high-resolution, small-receptive-field features to detect objects at different scales, and thus alleviates the aforementioned contradictory requirement between the feature map resolution and receptive fields. To further increase feature map resolution while keeping the receptive field, DetNet~\cite{li2018detnet} employs dilated convolutions and adds an extra stage. Until now, FPN-based approaches \myblue{(\eg, FPN and DetNet)} have reached the state-of-the-art performance in object detection. Nevertheless, the receptive field of these models is still much smaller than their input size.

	\begin{table}[t]
		\centering
		\caption{Detection results using ResNet-101 FPN~\cite{lin2017feature} with different input image sizes on COCO {\qcr{minival}}.}
			\resizebox{1.0\linewidth}{!}
		{
			\begin{tabular}{c|cccccc}
				\toprule
				Image size & AP & $\text{AP}_{50}$ & $\text{AP}_{75}$ & $\text{AP}_S$ & $\text{AP}_M$ & $\text{AP}_L$ \\
				\hline
				$600\times1000$ & 37.9  & 59.5  & 41.2  & 19.8 & 41.3  & 51.6 \\
				$800\times1333$ & 39.4  & 61.2  & 43.4  & 22.5 & 42.9  & 51.3 \\
				$1000\times1433$ & 39.5  & 61.6  & 43.0  & 24.0 & 43.0  & 49.6 \\
				\bottomrule
			\end{tabular}%
		}
		\label{tab:impact_image_size}%
	\end{table}%

	
	In addition, due to the limitation of the network architecture, FPN-based approaches cannot make good use of the receptive fields of different sizes. Specifically, the bottom-up pathway simply stacks layers to enlarge the receptive field without encouraging information propagation, and the feature maps corresponding to different receptive fields are just merged by element-wise addition in the top-down pathway. 
	Therefore, semantic information captured by different receptive fields does not well in communicating with each other, leading to the limited performance.

	\myblue{In short, there exist two main problems in current FPN-based approaches: 1) the contradictory requirement between feature map resolution and receptive  field on high-resolution inputs, and 2) the lack of effective communication among multi-size receptive fields.}
	To effectively tackle these two problems, we propose a module, called \textbf{Context Extraction Module (CEM)}.
	Without significantly increasing the computational overhead, CEM can capture rich context information from different large receptive fields by using multi-path dilated convolutional layers with different dilation rates (Fig.~\ref{fig:samples} (c)).
	Furthermore, to merge multi-receptive-field information elaborately, we introduce dense connections between the layers \myblue{with} different receptive fields in CEM.
	
	Nevertheless, although the feature from CEM contains rich context information and substantially helps to detect objects of different scales, we found that it is somewhat miscellaneous and thereby might confuse the localization and recognition tasks. Thus, as shown in Fig.~\ref{fig:samples} (d), to reduce the misleading of redundant context and further enhance the discriminative ability of feature, we design another module named \textbf{Attention-guided Module (AM)}, which introduces a self-attention mechanism to capture effective contextual dependencies. Specifically, it consists of two parts: 1) \textit{Context Attention Module (CxAM)} which aims at capturing semantic relationship between any two positions of the feature maps, and 2) \textit{Content Attention Module (CnAM)} which focuses on discovering spatial dependencies.
	
	In this paper, we name our whole model, which consists of CEM and AM, as \textbf{Attention-guided Context Feature Pyramid Network (AC-FPN)}. Our proposed AC-FPN can readily be plugged into existing FPN-based \myblue{model} and be easily trained end-to-end without additional supervision.

	We compare our AC-FPN with several state-of-the-art baseline methods on the COCO dataset. Extensive experiments demonstrate that, without any bells and whistles, our model achieves the best performance. Embedding the baselines with our modules (CEM and AM) significantly improves the performance on object detection. 
    \myblue{Furthermore, we also validate the proposed method on the more challenging instance segmentation task, and the experimental results show that the models integrated with CEM and AM substantially outperform the counterparts (\ie, without them).}
	The source code will be made publicly available.

	We highlight our principal contributions as follows:
	\begin{itemize}
		\item To address the contradictory requirement between feature map resolution and receptive fields in high-resolution images, we design a module named CEM to leverage features from multiple large contexts.
		\item In addition to producing more salient context information and further enhance the discriminative ability of feature representations, we introduce an attention-guided module named AM.
		\item Our two modules (CEM and AM, named together as AC-FPN) can readily be plugged into existing FPN-based models, \eg, PANet~\cite{liu2018path}, and 
		\myblue{significantly boost the performance in both object detection and instance segmentation tasks.}
	\end{itemize}

	\begin{figure*}[t]
		\begin{center}
			\includegraphics[width=1.0\linewidth]{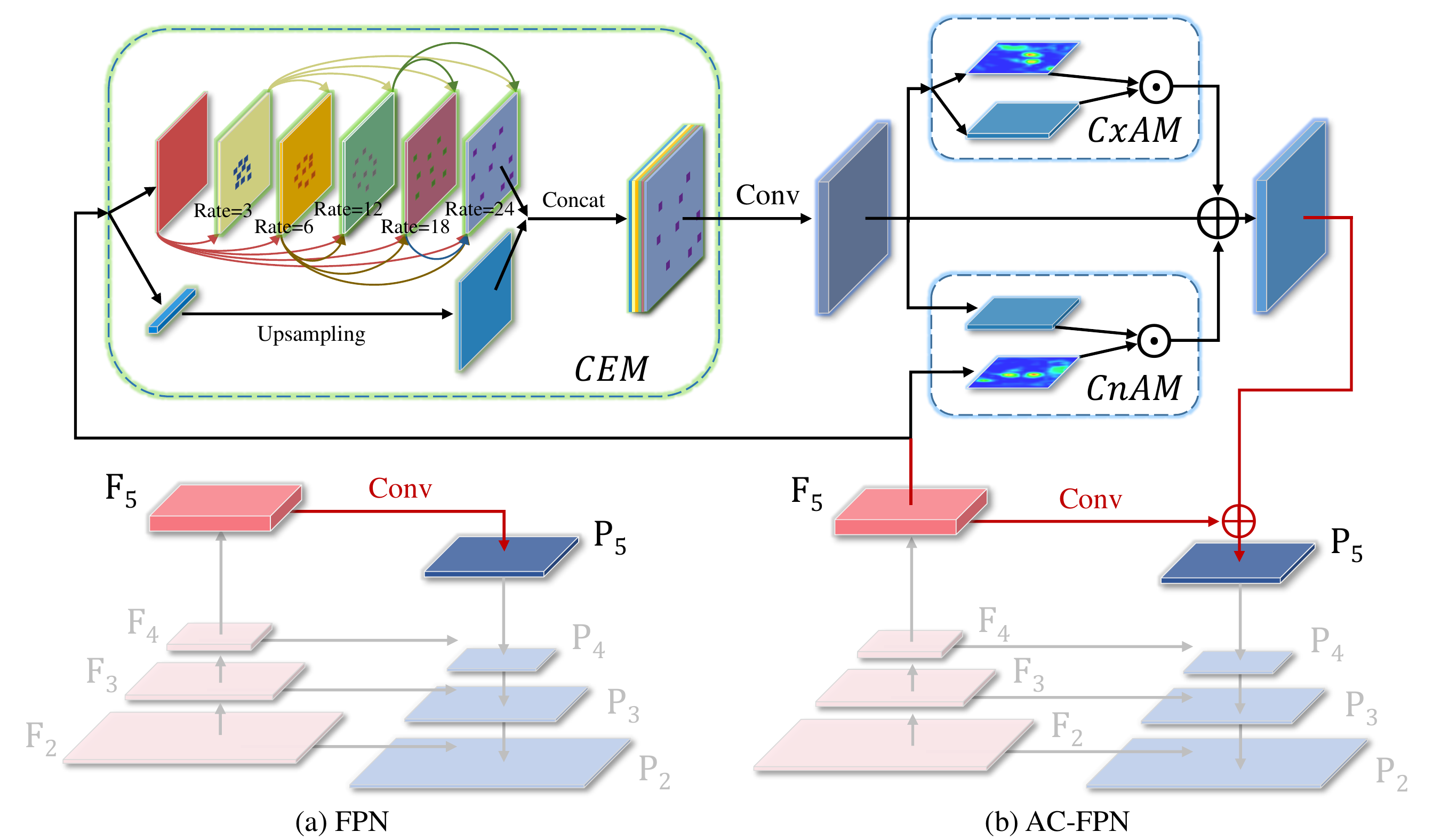}
		\end{center}
		\caption{The overall architecture of our proposed modules. Based on the structure of FPN, Context Extraction Module (CEM) is trained to capture the rich context information for various receptive fields and then produces an integrated representation. The Context Attention Module (CxAM) and Content Attention Module (CnAM) are devised to identify the salient dependencies among the extracted context.}
		\label{fig:architecture1}
	\end{figure*}

	\section{Related Work}
	
	\subsection{Object Detection}
	
	
	The frameworks of object detection in deep learning can 
    \myblue{be mainly divided into two categories:}
    1) two-stage detectors and 2) one-stage detectors.
	The \myblue{two-stage detectors} generate \myblue{thousands of} region proposals and then classifies each proposal into different object categories. For example, R-CNN~\cite{girshick2014rich} generates $2,000$ candidate proposals by Selective Search~\cite{uijlings2013selective} while filtering out the majority of negative positions \myblue{in the first stage}, and classifies the previous proposals into different object categories \myblue{in the second stage}.
	\myblue{Afterward}, Ross Girshick proposes Fast R-CNN~\cite{girshick2015fast}, which shares the convolution operations \myblue{and thus enables end-to-end training of classifiers and regressors.}
	Moreover, for Faster R-CNN~\cite{ren2015faster}, Region Proposal Networks (RPN) integrates proposal generation with the classifier into a single convolutional network.
	Numerous extensions of this framework have been proposed, such as R-FCN~\cite{dai2016r}, FPN~\cite{lin2017feature}, Mask R-CNN~\cite{he2017mask}, Cascade R-CNN~\cite{cai2018cascade}, CBNet~\cite{liu2019cbnet} and DetNet~\cite{li2018detnet}.
	
	The other regards object detection as a regression or classification problem, adopting a unified network to achieve final results (locations and categories) directly. 
	OverFeat~\cite{sermanet2013overfeat} is one of the first models with the one-stage framework on deep networks. Afterward, Redmon \etal propose YOLO~\cite{redmon2016you} \myblue{to make} use of the whole topmost feature map to predict both \myblue{classification confidences} and bounding boxes. In addition, Liu \etal devise SSD~\cite{liu2016ssd} to handle objects of various sizes using multi-scale bounding boxes on multiple feature maps. 
	\myblue{Besides, there are extensive other one-stage models enhancing the detection process in the prediction objectives or the network architectures, such as}
	YOLOv2~\cite{redmon2017yolo9000}, DSSD~\cite{fu2017dssd} and DSOD~\cite{shen2017dsod}.
	
	\subsection{Context Information}
	
	Context information can facilitate the performance of localizing the region proposals and thereby improve the final results of detection and classification. According to that, Bell \etal present Inside-Outside Net (ION)~\cite{bell2016inside} that exploits information both inside and outside the regions of interest. Chen \etal propose a context refinement algorithm~\cite{chen2018context}, which explores rich contextual information to better refine each region proposals. Besides, Hu \etal design a relation model~\cite{hu2018relation} that focuses on the interactions between each object that can be regarded as a kind of contextual cues. \myblue{Unlike} the plain architecture in~\cite{hu2018relation}, the models in~\cite{chen2017spatial,li2017attentive,stewart2016end} consider the relations with a sequential fashion. Moreover, for reasoning the obstructed objects, Chen \etal present a framework~\cite{chen2018iterative} to exploit the relational and contextual information by knowledge graphs.
	
	\subsection{Attention Modules}
	
	Attention modules can model long-range dependencies and become the workhorse of many challenging tasks, including image classification~\cite{mnih2014recurrent,wang2017residual}, semantic and instance segmentation~\cite{chen2016attention,ren2017end}, image captioning~\cite{lu2017knowing,xu2015show,you2016image}, natural language processing~\cite{bahdanau2014neural,lin2017structured,tan2018deep,vaswani2017attention}, \etc.
	For object detection, Li \etal propose a MAD unit~\cite{li2018zoom} to aggressively search for neuron activations among feature maps from \myblue{both} low-level and high-level streams. Likewise, to improve the detection performance, Zhu \etal design an Attention CoupleNet~\cite{zhu2019attention} that incorporates the attention-related information \myblue{with} global and local properties of the objects. Moreover, Pirinen \etal present a drl-RPN~\cite{pirinen2018deep} that replaces the typical RoI selection process of RPN~\cite{ren2015faster} with a sequential attention mechanism, which is optimized via deep reinforcement learning (RL).
	
	\section{Proposed Method}

	
	\myblue{Recently, the hierarchical detection approaches like FPN~\cite{lin2017feature} and DetNet~\cite{li2018detnet} have achieved promising performance.}
	However, for larger input images, these models have to 
	\myblue{stack more convolutional layers}
	to ensure the appropriateness of receptive fields. Otherwise, they will be in a dilemma between feature map resolution and receptive fields.
	Besides, for these models, the representation ability of generated features are limited due to the lack of effective communication between \myblue{receptive fields with different sizes.}
	
	To alleviate these limitations, we propose a novel \textit{Attention-guided Context Feature Pyramid Network (AC-FCN)} that captures context information from  receptive fields \myblue{with different sizes} and produces the objective features with stronger discriminative ability. As shown in Fig.~\ref{fig:architecture1}, built upon the basic FPN architecture~\cite{lin2017feature}, our proposed model has two novel components: 1) \textit{Context Extraction Module (CEM)} that 
	exploits rich context information from receptive fields \myblue{with various sizes};
    2) \textit{Attention-guided Module (AM)} that enhances salient context dependencies.
	We will depict each part of our model in the following subsections.
	
	\subsection{Context Extraction Module}
	
	To integrate the contextual information from different receptive fields, we build the Context Extraction Module (CEM), which only contains several additional layers.
	To be specific, \myblue{as shown in} Fig.~\ref{fig:architecture1}, for the bottom-up pathway, we denote the output of the convolutional layer in each scale as \{$\mathbf{F}_2, \mathbf{F}_3, \mathbf{F}_4, \mathbf{F}_5$\} according to the settings in~\cite{lin2017feature}. Likewise, both the top-down pathway and lateral connections follow the official settings in the original paper~\cite{lin2017feature}.
	
	After obtaining the feature maps from preceding layers (\ie, $\mathbf{F}_5$), to exploit rich contextual information, we feed it into our CEM, which consists of multi-path dilated convolutional layers~\cite{yu2015multi} with different rates, \eg, $\text{rate}=3, 6, 12$.
	These separated convolutional layers can harvest multiple feature maps in various receptive fields.
	Besides, to enhance the capacity of modeling geometric transformations, we introduce deformable convolutional layers~\cite{dai2017deformable} in each path. It ensures our CEM can learn transformation-invariant features from the given data.


	In addition, to merge multi-scale information elaborately, we employ dense connections in our CEM, where the output of each dilated layer is concatenated with the input feature maps and then fed into the next dilated layer.
	DenseNet~\cite{huang2017densely} employs the dense connection to tackle the issues of 
    \myblue{vanishing gradients and strengthens feature propagation}
	when the CNN model is increasingly deep. 
    \myblue{By contrast,}
	we use the dense fashion to \myblue{achieve} better scale diversities of the features with various receptive fields. 
    Finally, \myblue{in order to maintain the coarse-grained information of the initial inputs, we concatenate the outputs of the dilated layers with the up-sampled inputs and feed them into a $1 \times 1$ convolutional layer to fuse the coarse-and-fine-grained features.}
	
	
	\subsection{Attention-guided Module}
	
	\begin{figure}[t]
		\begin{center}
\includegraphics[width=1.0\linewidth]{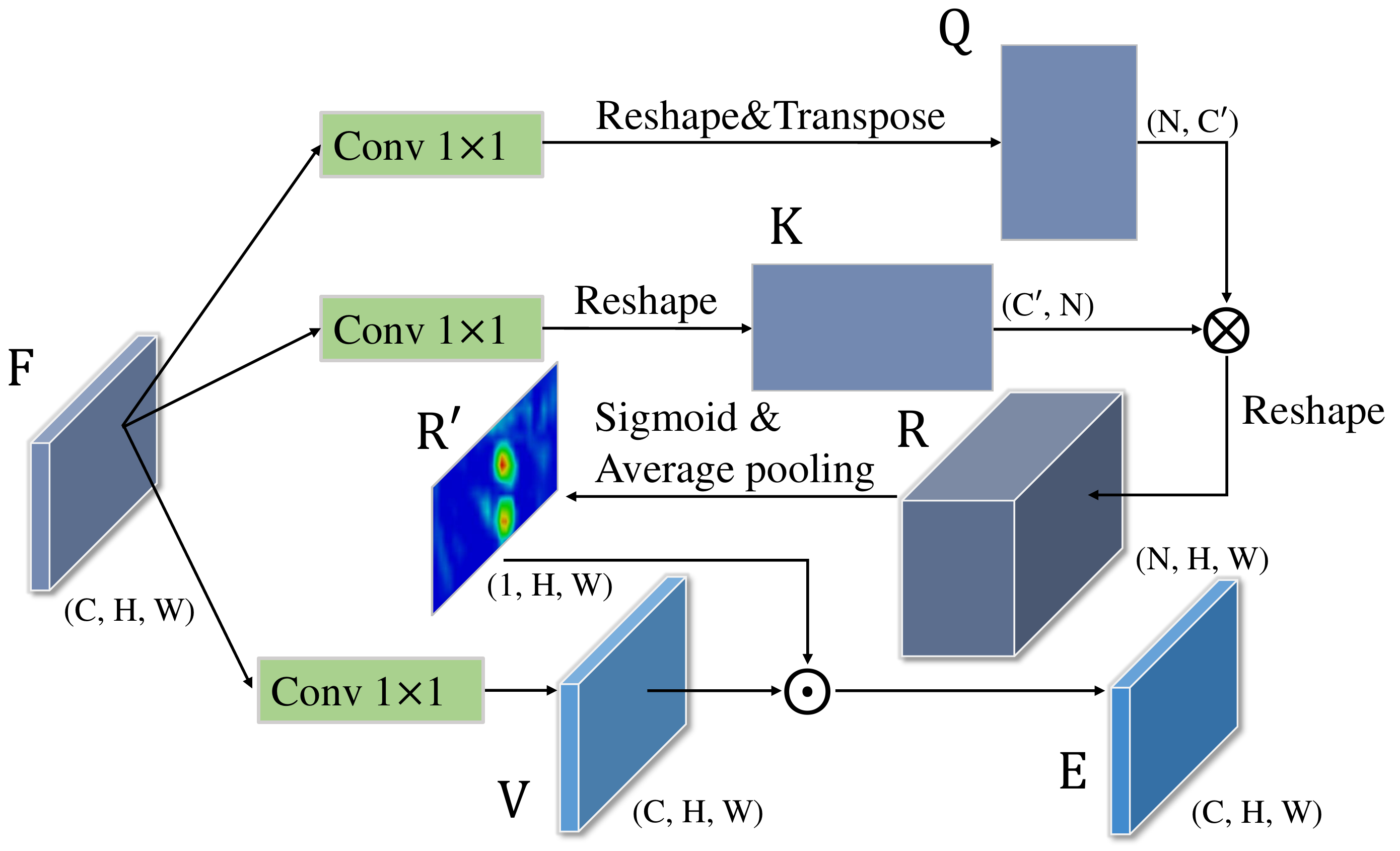}
		\end{center}
		\caption{Architecture of Context Attention Module (CxAM).}
		\label{fig:CxAM}
	\end{figure}

	\myblue{Although} the features from CEM contain rich receptive field information, not all of them are useful to facilitate the performance of object detection. The accuracy may reduce since the bounding boxes or region proposals are mislead by redundant information. Thus, to remove the negative impacts of the redundancy and further enhance the representation ability of feature maps, we propose an Attention-guided Module (AM), which is able to capture salient dependencies with strong semantics and precise locations. As shown in Fig.~\ref{fig:architecture1}, the attention module consists of two parts: 1) Context Attention Module (CxAM) and 2) Content Attention Module (CnAM).

	More specifically, CxAM focuses on the semantics between subregions \myblue{of} given feature maps (\ie, the features from CEM). However, due to the effects of the deformable convolution, the 
    location
	of each object has been destroyed dramatically. To alleviate this issue, we introduce CnAM, which pays more attention to ensure the spatial information but sacrifices some semantics due to the attention from the shallower layer (\ie, $\mathbf{F}_5$). 
    \myblue{Finally, the features refined by CxAM and CnAM are merged with the input features to obtain more comprehensive representations.}
	
	\subsubsection{Context Attention Modules}\label{CxAM}

	\begin{figure}[t]
		\begin{center}
\includegraphics[width=1.0\linewidth]{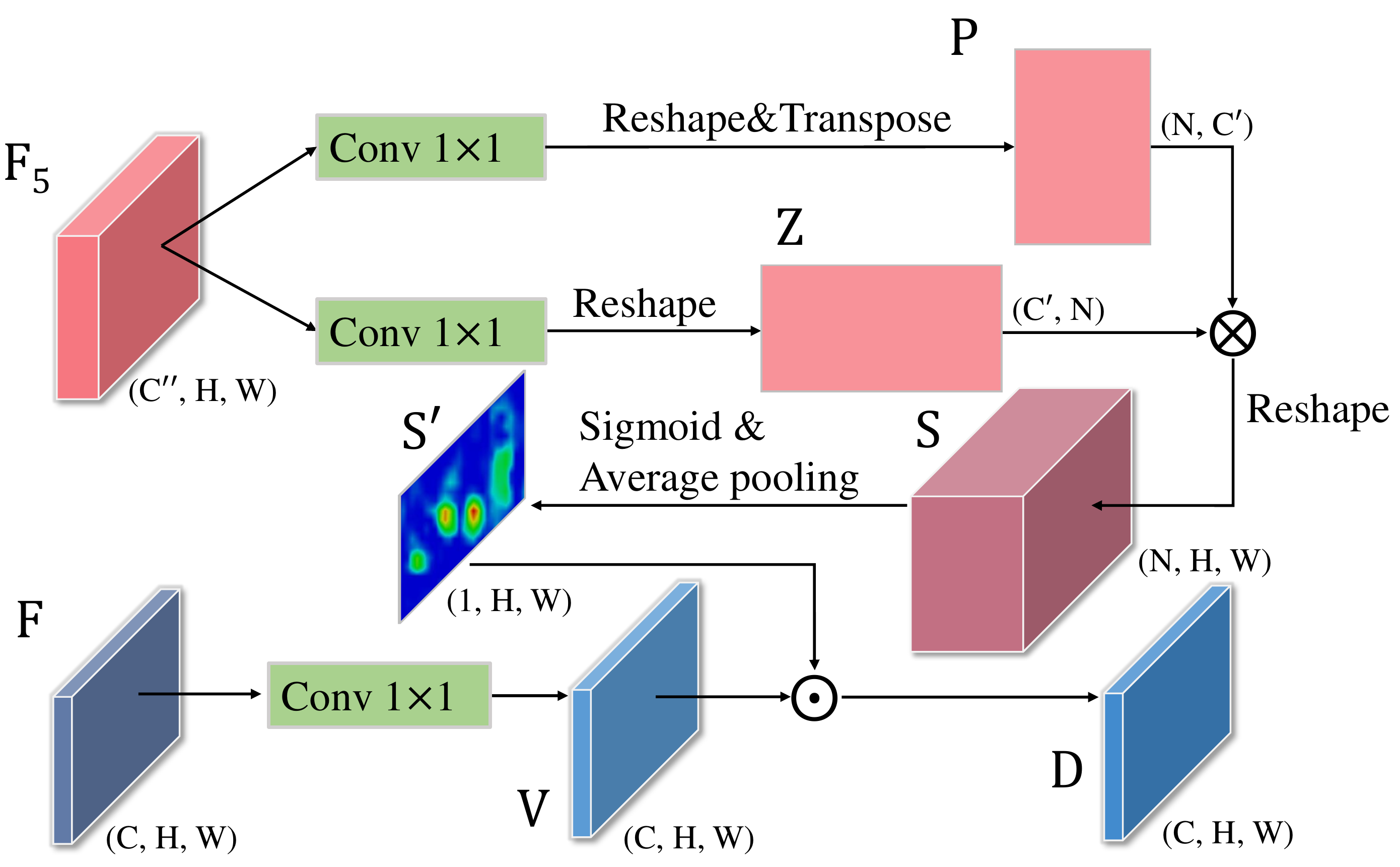}
		\end{center}
		\caption{Architecture of Content Attention Module (CnAM).}
		\label{fig:CnAM}
	\end{figure}

	To actively capture the semantic dependencies between subregions,
	we introduce a Context Attention Module (CxAM) based on the self-attention mechanism. \myblue{Unlike} \cite{fu2018dual}, we feed the preceding features, which are produced by CEM and contain multi-scale receptive field information, into CxAM module. Based on these informative features, CxAM adaptively pays more attention to the relations between subregions which are more relevant. Thus, the output features from CxAM will have clear semantics and contain contextual dependencies within surrounding objects.
	
	As can be seen in Fig.~\ref{fig:CxAM}, given discriminative feature maps $\mathbf{F}\in\mathbb{R}^{C\times H\times W}$, we transform them into a latent space by using the convolutional layers $\mathbf{W}_q$ and $\mathbf{W}_k$, respectively. The converted feature maps are calculated by 
	{
		\begin{equation}
		\mathbf{Q}={\mathbf{W}}^\top_q \mathbf{F}~~\text{and}~~ \mathbf{K}={\mathbf{W}}^\top_k \mathbf{F},
		\end{equation}}where $\{\mathbf{Q},\mathbf{K}\}\in\mathbb{R}^{C^\prime \times H\times W}$. Then, we reshape $\mathbf{Q}$ and $\mathbf{K}$ to $\mathbb{R}^{C^\prime \times N}$, where $N=H\times W$. To capture the relationship between each subregion, we calculate a \myblue{correlation} matrix as
	{
		\begin{equation}\label{R}
		\mathbf{R} = \mathbf{Q}^\top \mathbf{K},
		\end{equation}}where $\mathbf{R}\in\mathbb{R}^{N \times N}$ \myblue{and then be reshaped to $\mathbf{R}\in\mathbb{R}^{N \times H \times W}$.}
		\myblue{After} normalizing $\mathbf{R}$ via sigmoid activation function and average pooling, we build an attention matrix $\mathbf{R}^\prime$, where $\mathbf{R}^\prime\in\mathbb{R}^{1 \times H \times W}$.
	
	Meanwhile, we \myblue{transform} the feature map $\mathbf{F}$ to another representation $\mathbf{V}$ by using the convolutional layer $\mathbf{W}_v$:
	{
		\begin{equation}
		\mathbf{V}={\mathbf{W}}^\top_v \mathbf{F},
		\end{equation}}where $\mathbf{V}\in\mathbb{R}^{C\times H\times W}$.
	
	Finally, 
    \myblue{an element-wise multiplication is performed on $\mathbf{R}^\prime$ and the feature $\mathbf{V}$}
	to get the attentional representation $\mathbf{E}$. We formulate the function as
	{
		\begin{equation}
		\mathbf{E}_i = \mathbf{R}^\prime \odot \mathbf{V}_i,
		\end{equation}}where $\mathbf{E}_i$ refers to the $i^{th}$ feature map along with the channel dimension $C$.
	

	\subsubsection{Content Attention Module}

	\begin{table*}[t]
		\centering
		\caption{Detailed model design of the proposed CEM.}
		\resizebox{1.0\linewidth}{!}
		{
			\begin{tabular}{c|c|c|c}
				\toprule
				Module & Module details & Input shape & Output shape \\
				\hline
				CEM\_3\_1x1 & Conv  & (2048, w, h) & (512, w, h) \\
				\hline
				CEM\_3\_3x3 & DeformConv(dilate=3) & (512, w, h) & (256, w, h) \\
				\hline
				CEM\_concat\_1 & Concatenation(C5, CEM\_3\_3x3) & (2048, w, h)$\oplus$(256, w, h) & (2304, w, h) \\
				\hline
				CEM\_6\_1x1 & Conv  & (2304, w, h) & (512, w, h) \\
				\hline
				CEM\_6\_3x3 & DeformConv(dilate=6) & (512, w, h) & (256, w, h) \\
				\hline
				CEM\_concat\_2 & Concatenation(CEM\_concat\_1, CEM\_6\_3x3) & (2304, w, h)$\oplus$(256, w, h) & (2560, w, h) \\
				\hline
				CEM\_12\_1x1 & Conv  & (2560, w, h) & (512, w, h) \\
				\hline
				CEM\_12\_3x3 & DeformConv(dilate=12) & (512, w, h) & (256, w, h) \\
				\hline
				CEM\_concat\_3 & Concatenation(CEM\_concat\_2, CEM\_12\_3x3) & (2560, w, h)$\oplus$(256, w, h) & (2816, w, h) \\
				\hline
				CEM\_18\_1x1 & Conv  & (2816, w, h) & (512, w, h) \\
				\hline
				CEM\_18\_3x3 & DeformConv(dilate=18) & (512, w, h) & (256, w, h) \\
				\hline
				CEM\_concat\_4 & Concatenation(CEM\_concat\_3, CEM\_18\_3x3) & (2816, w, h)$\oplus$(256, w, h) & (3072, w, h) \\
				\hline
				CEM\_24\_1x1 & Conv  & (3072, w, h) & (512, w, h) \\
				\hline
				CEM\_24\_3x3 & DeformConv(dilate=24) & (512, w, h) & (256, w, h) \\
				\hline
				CEM\_global\_context & Global Averge Pooling & (2048, w, h) & (2048, 1, 1) \\
				\hline
				CEM\_gc\_reduce\_1x1 & Conv  & (2048, 1, 1) & (256, 1, 1) \\
				\hline
				CEM\_gc\_upsample & Bilinear Interpolation & (256, 1, 1) & (256, w, h) \\
				\hline
				CEM\_concat\_5 & \tabincell{c}{Concatenation(CEM\_3\_3x3,  CEM\_6\_3x3,  CEM\_12\_3x3, \\ CEM\_18\_3x3, CEM\_24\_3x3, CEM\_gc\_upsample)} & \tabincell{c}{(256, w, h)$\oplus$(256, w, h)$\oplus$(256, w, h)$\oplus$\\ (256, w, h)$\oplus$(256, w, h)$\oplus$(256, w, h)} & (1536, w, h) \\
				\hline
				CEM\_reduce\_1x1 & Conv  & (1536, w, h) & (256, w, h) \\
				\bottomrule
			\end{tabular}%
		}
		\label{tab:details_cem}%
	\end{table*}%

	Due to the effects of deformable convolutions in CEM, the geometric properties of the given images have been destroyed drastically, leading to the \myblue{location offsets.} To solve this problem, we design a new attention module, called Content Attention Module (CnAM), to maintain precise \myblue{position information} of each object.
	
	
	As shown in Fig.~\ref{fig:CnAM}, similar to CxAM, we use convolutional layers to transform the given feature maps. However, instead of using the feature maps $\mathbf{F}$ to produce the attention matrix, we adopt the feature maps $\mathbf{F}_5\in\mathbb{R}^{C^{\prime\prime}\times H\times W}$, which can capture the more precise location
	of each object.
	
	To get the attention matrix, at first we apply two convolutional layers $\mathbf{W}_p$ and $\mathbf{W}_z$, to convert $\mathbf{F}_5$ into the latent space, respectively:
	{
		\begin{equation}
		\mathbf{P} = {\mathbf{W}}^\top_p \mathbf{F}_5~~\text{and}~~\mathbf{Z} = {\mathbf{W}}^\top_z \mathbf{F}_5,
		\end{equation}}where $\{\mathbf{P},\mathbf{Z}\}\in\mathbb{R}^{C^\prime \times H\times W}$. Then, we reshape  the dimension of $\mathbf{P}$ and $\mathbf{Z}$ to $\mathbb{R}^{C^\prime \times N}$, \myblue{and} produce the \myblue{correlation} matrix similar to Eq.~(\ref{R}) as:
	{
		\begin{equation}
		\mathbf{S} = \mathbf{P}^\top \mathbf{Z},
		\end{equation}}where $\mathbf{S}\in\mathbb{R}^{N \times N}$. 
	After \myblue{reshaping} $\mathbf{S}$ to $\mathbb{R}^{N \times H \times W}$, we employ sigmoid function and average pooling to produce an attention matrix $\mathbf{S}^\prime \in \mathbb{R}^{1 \times H \times W}$.
	To obtain a prominent representation, we combine the extracted features $\mathbf{V}$ (see Section~\ref{CxAM}) with $\mathbf{S}^\prime$ by element-wise multiplication:
	{
		\begin{equation}
		\mathbf{D}_i = \mathbf{S}^\prime \odot \mathbf{V}_i,
		\end{equation}}where $\mathbf{D} \in \mathbb{R}^{C \times H \times W}$ and $\mathbf{D}_i$ indicates the $i^{th}$ \myblue{output feature map.}

	\section{Experiments on Object Detection}

	In this section, we evaluate the performance of our AC-FPN compared to the baseline methods including Cascade R-CNN~\cite{cai2018cascade}, Faster R-CNN with FPN~\cite{lin2017feature},
	PANet~\cite{liu2018path} and DetNet~\cite{li2018detnet}.
	Following the settings in~\cite{cai2018cascade,liu2018path}, we train our model on MS-COCO 2017~\cite{lin2014microsoft}, which consists of 115k training images and 5k validation images ({\qcr{minival}}). We also report the final results on a set of 20k test images ({\qcr{test-dev}}). For quantitative comparison, we use the COCO-style Average Precision (AP) and PASCAL-style AP (\ie averaged over IoU thresholds). Following the definitions in~\cite{lin2014microsoft}, on COCO, we denote the objects with small, medium and large sizes as $\text{AP}_S$, $\text{AP}_M$ and $\text{AP}_L$, respectively. For PASCAL-style AP, we use $\text{AP}_{50}$ and $\text{AP}_{75}$, which are defined at a single IoU of 0.5 and 0.75.

	\subsection{Implementation Details.}

	Following the settings in~\cite{cai2018cascade,he2017mask,liu2018path}, we resize the input images with the \myblue{shorter side of} $800$ pixels and initialize the \myblue{feature extractor with a model pretrained on ImageNet.}
	Specifically, we train our model with learning rate $0.02$ for 60k iterations and reduce to $0.002$ for another 20k iterations. 
	For a fair comparison, we train our model without any data augmentations \myblue{except the} horizontal image flipping.
	
	For our AC-FPN, different from the original FPN, we use dilated convolution on $\mathbf{F}_5$ and subsample $\mathbf{P}_5$ via max pooling to keep the same stride with FPN. More specifically, in CEM, we first reduce $\mathbf{F}_5$ to $512$ channels as input, followed by several $3 \times 3$ deformable convolutional layers with different dilated rates, \eg, $3, 6, 9$. Then we reduce the output to $256$ channels \myblue{for reusing} the top-down structure of FPN.
	\qi{More details are shown in Tab.~\ref{tab:details_cem}.
	In AM, as shown in Figs.~\ref{fig:CxAM} and~\ref{fig:CnAM}, we use $1 \times 1$ convolution to reduce the input to $256$ channels in CnAM and $128$ channels in CxAM, respectively.} For compared methods, 
	we use the reimplementations and settings in Detectron~\cite{girshick2018detectron}. 


	\begin{table*}[t]
		\centering
		\caption{The state-of-the-art 
			detectors on COCO {\qcr{test-dev}}. The entries denoted by ``*" use the implementations of Detectron\cite{girshick2018detectron}. ``AC-Cascade" means Cascade R-CNN embedded with AC-FPN.}
		{
			\begin{tabular}{c|c|cccccc}
				\toprule
				Methods &  Backbone & AP  & $\text{AP}_{50}$ & $\text{AP}_{75}$ & $\text{AP}_{S}$ & $\text{AP}_{M}$ & $\text{AP}_{L}$ \\
				\hline
				FPN*~\cite{lin2017feature} & ResNet-50 & 37.2 & 59.3  & 40.2  &  20.9 & 39.4  & 46.9  \\
				FPN & DetNet-59~\cite{li2018detnet} & 40.3 & 62.1 & 43.8 & 23.6 & 42.6 & 50.0  \\
				\hline
				FPN*~\cite{lin2017feature} & ResNet-101 & 39.4 & 61.5  &  42.8  & 22.7 & 42.1 & 49.9 \\
				DRFCN~\cite{dai2017deformable} & ResNet-101 & 37.1 & 58.9 & 39.8 & 17.1 & 40.3 & 51.3  \\     
				Mask R-CNN*~\cite{he2017mask} & ResNet-101 &  40.2 & 62.0 & 43.9 & 22.8   & 43.0 & 51.1  \\
				Cascade R-CNN*~\cite{cai2018cascade} &  ResNet-101 & 42.9 & 61.5 & 46.6  & 23.7  & 45.3 &  55.2 \\
				C-Mask R-CNN~\cite{chen2018context} & ResNet-101 & 42.0 & 62.9 & 46.4 & 23.4 & 44.7 & 53.8 \\
				\hline
				AC-FPN* & ResNet-50 &  40.4 & 63.0 & 44.0 & 23.5 & 43.0 & 50.9  \\
				AC-FPN* & ResNet-101 & 42.4  & \textbf{65.1} & 46.2 & 25.0 & 45.2 & 53.2  \\
				AC-Cascade* & ResNet-101 & \textbf{45.0}  & 64.4 & \textbf{49.0}  &  \textbf{26.9} &  \textbf{47.7} & \textbf{56.6} \\
				\bottomrule
		\end{tabular}}
		\label{tab:state-of-the-art}%
	\end{table*}%

	\begin{table*}[t]
		\centering
		\caption{Detailed comparisons on multiple popular baseline object detectors on the COCO dataset. We plug our modules (CEM and AM) into the existing methods and test their results based on both ResNet-50 and ResNet-101.}
		\resizebox{1.0\linewidth}{!}
		{
			\begin{tabular}{c|c|c|cccccc|cccccc}
				\toprule
				\multirow{2}{*}{Methods} & \multirow{2}{*}{Backbone} & \multicolumn{1}{r|}{\multirow{2}{*}{ + Our modules}} & \multicolumn{6}{c|}{{\qcr{minival}}}                 & \multicolumn{6}{c}{{\qcr{test-dev}}} \\
				\cline{4-15}          &       &       & AP    & $\text{AP}_{50}$ & $\text{AP}_{75}$ & $\text{AP}_{S}$ & $\text{AP}_{M}$ & $\text{AP}_{L}$ & AP   & $\text{AP}_{50}$ & $\text{AP}_{75}$ & $\text{AP}_{S}$ & $\text{AP}_{M}$ & $\text{AP}_{L}$ \\
				\hline
				\multirow{4}{*}{ 
					FPN~\cite{lin2017feature}} & \multirow{2}[2]{*}{ResNet-50} &       & 36.7  & 58.4  & 39.6  & 21.1  & 39.8  & 48.1  & 37.2 & 59.3  & 40.2  &  20.9 & 39.4  & 46.9 \\
				&       & $\surd$     & 40.1 & 62.5 & 43.2 & 23.9 & 43.6 & 52.4 & 40.4 & 63.0 & 44.0 & 23.5 & 43.0 & 50.9\\
				\cline{2-15}          & \multirow{2}[2]{*}{ResNet-101} &       & 39.4 & 61.2 & 43.4 & 22.6 & 42.9  & 51.4 & 39.4 & 61.5  &  42.8  & 22.7 & 42.1 & 49.9  \\
				&       & $\surd$     & 42.0 & 64.7 & 45.6 & 25.1 & 45.7 & 53.4 & 42.4  & 65.1 & 46.2 & 25.0 & 45.2 & 53.2\\
				\hline
				\multirow{4}{*}{PANet~\cite{liu2018path}} & \multirow{2}[2]{*}{ResNet-50} &       & 38.7 & 60.4 & 41.7 & 22.6 & 42.4 & 50.3 & 39.0 & 60.8 & 42.1 & 22.2 & 41.7 & 48.7 \\
				&       & $\surd$     & 40.8 & 62.4 & 44.3 & 24.1 & 44.7 & 53.0  & 40.9 & 62.8 & 44.3 & 23.6 & 43.6 &  51.6 \\
				\cline{2-15}          & \multirow{2}[2]{*}{ResNet-101} &       & 40.5 & 62.0 & 43.8 & 23.0 & 44.8 & 53.2 & 40.8 & 62.7 & 44.2 & 23.2 & 43.9 & 51.7\\
				&       & $\surd$     & 42.7 & 64.4 & 46.5 & 25.5 & 46.7 & 54.9 & 43.0 & 65.1 & 46.8 & 25.6 & 46.1 & 53.6 \\
				\hline
				\multirow{4}{*}{Cascade R-CNN~\cite{cai2018cascade}} & \multirow{2}[2]{*}{ResNet-50} &       & 40.9 & 59.0 & 44.6 & 22.5 & 43.6 & 55.3 & 41.1  & 59.6  &  44.6 & 22.8  & 43.0  & 53.2  \\
				&       & $\surd$     & 43.0 & 62.4 & 46.7 & 25.3 & 46.4  & 56.4 & 43.3 & 62.7 & 47.4 & 25.0 & 45.7 & 55.1 \\
				\cline{2-15}          & \multirow{2}[2]{*}{ResNet-101} &       & 42.8 & 61.4 & 46.8 & 24.1 & 45.8 & 57.4 & 42.9 & 61.5 & 46.6  & 23.7  & 45.3 &  55.2\\
				&       & $\surd$     & 44.9 & 64.3 & 48.7 & 27.5 & 48.7 & 57.8 & 45.0 & 64.4 & 49.0 &  26.9 &  47.7 & 56.6 \\
				\bottomrule
		\end{tabular}}
		\label{tab:object_detection}%
	\end{table*}%

	\subsection{Comparisons with State-of-the-arts}

	\subsubsection{Quantitative Evaluation}

	Tab.~\ref{tab:state-of-the-art} shows the \myblue{detection performance} of state-of-the-art methods \myblue{on COCO {\qcr{test-dev}}}. 
	Compared to FPN (ResNet-50) and FPN (DetNet-59), our AC-FPN (ResNet-50) consistently achieves the best performance. To be specific, compared to FPN (ResNet-50), the promotions of $\text{AP}_S$, $\text{AP}_M$ and $\text{AP}_L$ are $2.4$, $3.7$ and $4.1$, respectively. Likewise, for FPN (DetNet-59), 
	\myblue{we also obtain the biggest improvement in $\text{AP}_L$, which}
	demonstrates that our model is able to capture \myblue{much more effective} information from large receptive fields. 

	On the other hand, compared to FPN (ResNet-101), although it contains more layers and establishes a deeper network, the performance even can not surpass our AC-FPN's built upon ResNet-50. Hence, we can conjecture that even if both FPN (ResNet-101) and AC-FPN (ResNet-50) have large receptive fields, our improvements are still relatively large since \myblue{the proposed method extracts much} stronger semantics and context information.

	
	\subsubsection{Adaptation to Existing Methods}
	
	To investigate the effects of our proposed modules, we embed our CEM and AM into some existing models. For a fair comparison, we use the same hyper-parameters \myblue{for} both baseline models and ours. Besides, we train two models for each baseline with different backbones, \ie, ResNet-50 and ResNet-101. 
	Tab.~\ref{tab:object_detection} shows that the models with CEM and AM consistently outperform their counterparts without them. \myblue{And we argue that the} reasons lie in that our modules can capture much richer context information from various receptive fields.

	\subsubsection{Qualitative Evaluation}
	
\begin{figure*}[t]
		\centering
		{\scriptsize
			\resizebox{1.0\linewidth}{!}{
				\begin{tabular}{>{\centering\arraybackslash} m{4.5cm}
						>{\centering\arraybackslash} m{7.5cm}
						>{\centering\arraybackslash} m{8.9cm} >{\centering\arraybackslash} m{3.9cm} >{\centering\arraybackslash} m{10.6cm}
						>{\centering\arraybackslash} m{10.7cm}}
					\huge{FPN w/o ~~~ CEM and AM} &\includegraphics[height=60mm]{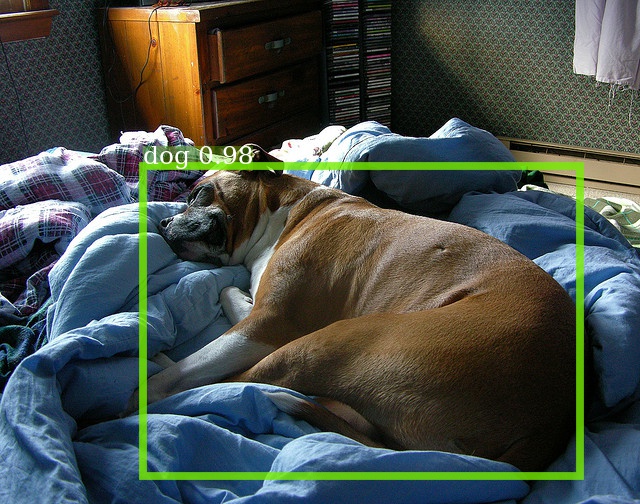}
					&\includegraphics[height=60mm]{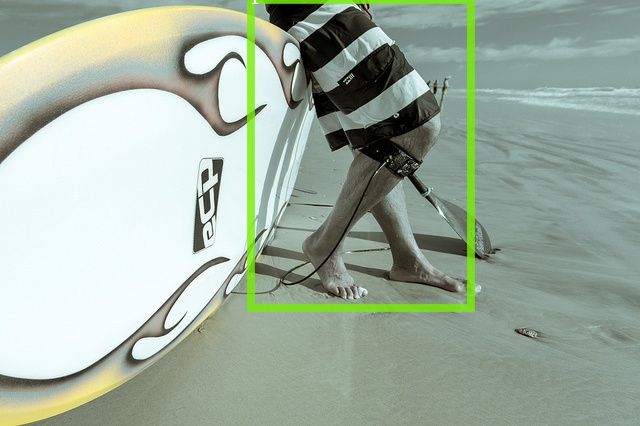}
					&\includegraphics[height=60mm]{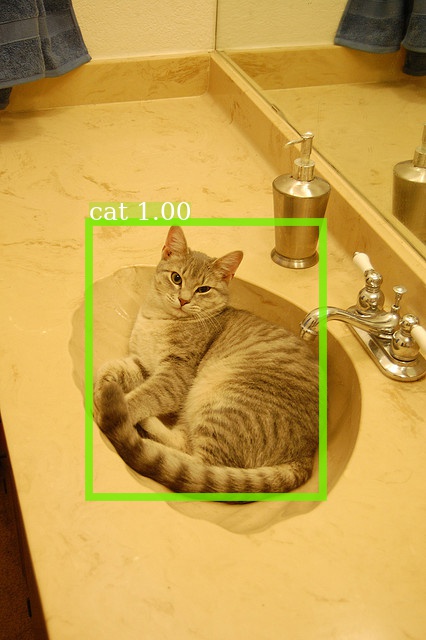}
					&\includegraphics[height=60mm]{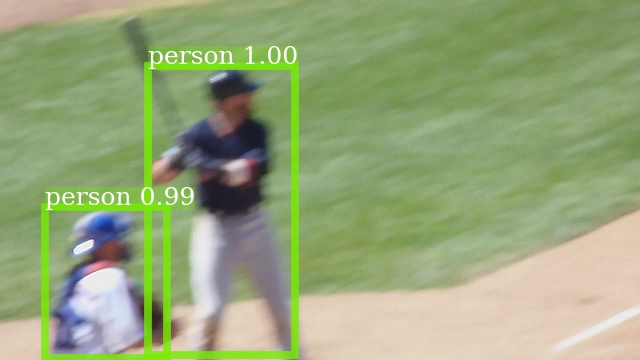}
					&\includegraphics[height=60mm]{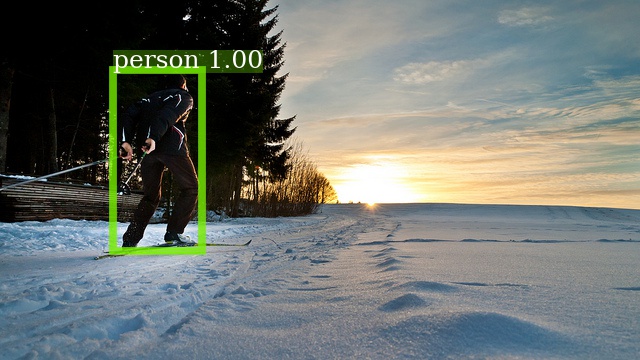}
					\\ \\
					\huge{FPN with ~~~ CEM and AM} &\includegraphics[height=60mm]{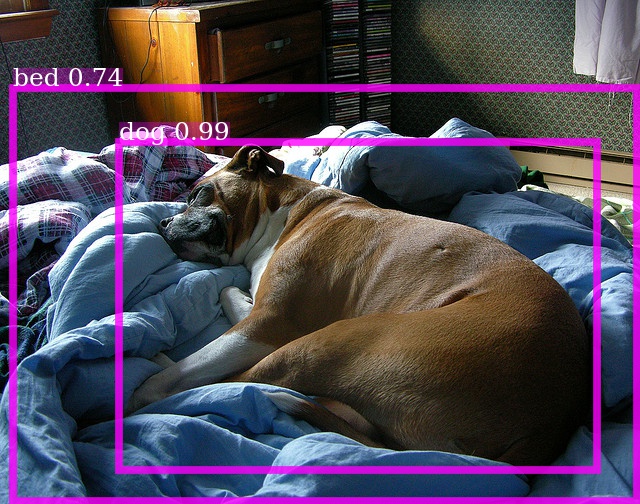}
					&\includegraphics[height=60mm]{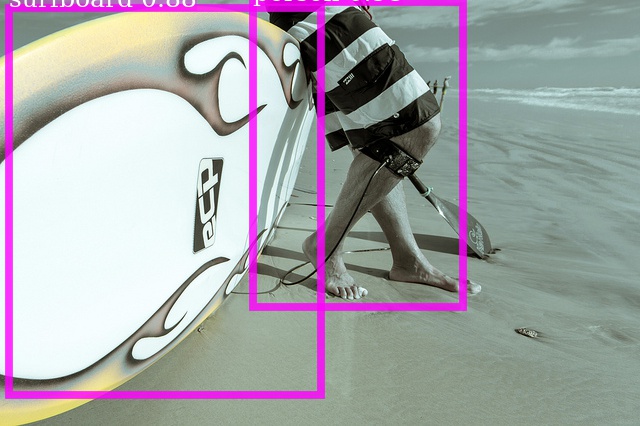}
					&\includegraphics[height=60mm]{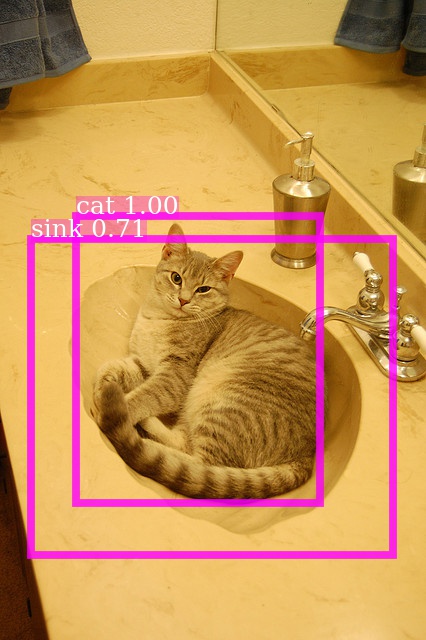}
					&\includegraphics[height=60mm]{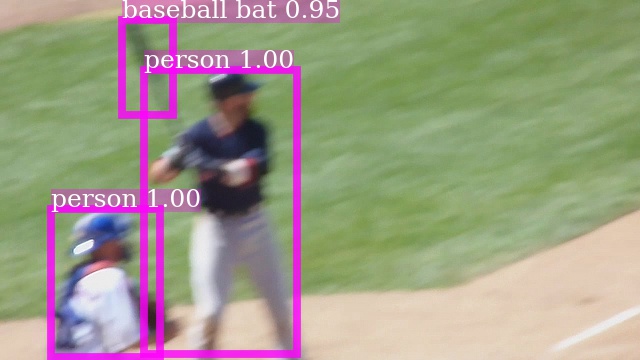}
					&\includegraphics[height=60mm]{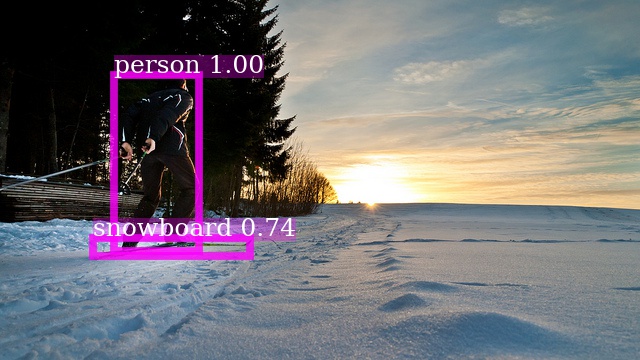}
					\\ \\
					& \Huge{(a)} & \Huge{(b)} & \Huge{(c)} & \Huge{(d)} & \Huge{(e)}
			\end{tabular}}
		}
		\caption{Visualization of object detection. Both models are built upon ResNet-50 on COCO {\qcr{minival}}.}
		\label{detection_visualization}
	\end{figure*}

	Moreover, we show the visual results of FPN with or without our CEM and AM. For a fair comparison, we build the models upon ResNet-50 and test on COCO {\qcr{minival}}. Besides, for convenient, we compare the detection performance of the same images with $\text{threshold}=0.7$. From Fig.~\ref{detection_visualization} (a), (b) and (c), the typical FPN model misses some large objects since these objects may be out of the receptive fields and FPN can not capture. By contrast, with our modules, the integrated FPN overcomes this limitation by accessing richer receptive fields.
	\myblue{As shown in Figs.~\ref{detection_visualization} (d) and (e), our models also perform much better on the ambiguous objects}
	by exploring the context information from various receptive fields.
	
	\subsection{Experiments of Context Extraction Module}\label{sec:cem}
	
	\subsubsection{Effectiveness of CEM}
	
	To discuss the effects of CEM, we plug CEM into the existing models and compare the performance with that without it. \myblue{As shown in Tab.~\ref{tab:effect_extraction}}, compared to the baselines, \myblue{the models with CEM achieves} more compelling performance. It demonstrates that our proposed CEM is able to facilitate \myblue{the object detection task by capturing richer} context information from \myblue{receptive fields with different sizes}.
	
	%

	\begin{table}[t]
		\centering
		\caption{Effects of our CEM with ResNet-50 on COCO {\qcr{minival}}.}
		\resizebox{1.0\linewidth}{!}{
			\begin{tabular}{c|c|cccccc}
				\toprule
				Methods & + CEM & AP  & $\text{AP}_{50}$ & $\text{AP}_{75}$ & $\text{AP}_{S}$ & $\text{AP}_{M}$ & $\text{AP}_{L}$ \\
				%
				\hline
				\multicolumn{1}{c|}{\multirow{2}[2]{*}{
						FPN}}
				
				&       &  36.7  & 58.4  & 39.6  & 21.1  & 39.8  & 48.1 \\
				& $\surd$     & 39.3 & 61.9 & 42.6 & 23.6 & 42.9 & 50.4 \\
				\hline
				\multicolumn{1}{c|}{\multirow{2}[2]{*}{PANet}}
				
				&       & 38.7 & 60.4 & 41.7 & 22.6 & 42.4 & 50.3 \\
				& $\surd$  & 39.9 & 62.3 & 43.3 & 23.6 & 43.6 & 52.0 \\
				\hline
				\multicolumn{1}{c|}{\multirow{2}[2]{*}{Cascade R-CNN}}
				
				&       & 40.9 & 59.0 & 44.6 & 22.5 & 43.6 & 55.3  \\
				& $\surd$     &  42.1 & 61.1 & 45.8 & 22.7 & 45.8 & 57.1 \\
				\bottomrule
		\end{tabular}}
		\label{tab:effect_extraction}%
	\end{table}%

	
	\subsubsection{Impacts of Deformable Convolutions}

	\qi{
		To \myblue{evaluate the impact} of deformable convolutions in our framework, we conduct an ablation study, where we compare the results of object detection using our proposed CEM module with and without deformable convolutions.
		Note that in order to \myblue{avoid interference from other modules}, we remove the proposed AM module from the whole framework.
		From Tab.~\ref{tab:deformable}, the results show that with deformable convolution,
		\myblue{the $\text{AP}_L$ improves while the $\text{AP}_S$ goes down slightly.}
		The reason may lie on that the deformable convolutions destroy the location information
		when seeking more powerful contextual information, especially for small objects, which are more sensitive to the locations.
	}

	\begin{table}[t]
		\centering
		\caption{Results of detection using resnet-50 backbone with and w/o deformable convolutions on COCO {\qcr{minival}}.}
		\resizebox{1.0\linewidth}{!}
		{
			\begin{tabular}{cc|cccccc}
				\toprule
				Deformable & Rest of CEM & AP & $\text{AP}_{50}$ & $\text{AP}_{75}$ & $\text{AP}_S$ & $\text{AP}_M$ & $\text{AP}_L$ \\
				\hline
				&  & 36.7 & 58.4 & 39.6 & 21.1  & 39.8 &  48.1 \\
				& $\surd$   &  38.8  & 61.8  & 41.8  & 23.7  & 42.5  & 49.7 \\
				$\surd$   &    $\surd$      & 39.2  & 61.9  & 42.5  & 23.5  & 42.9  & 50.3 \\
				\bottomrule
			\end{tabular}%
		}
		\label{tab:deformable}%
	\end{table}%

	\subsubsection{Effects of Dense Connections}
	
	\qi{
		To investigate the impact of dense connections in CEM, we report the detection results on COCO {\qcr{minival}} with or without the dense connections. \myblue{Similarly}, to reduce the influence of other factors, we remove the proposed AM module and deformable convolutions from the whole framework. As shown in Tab.~\ref{tab:dense_connection}, when using dense connection method, the performance increases consistently in all metrics, which demonstrates that the dense connections are effective and an elaborate fusion method has a positive impact in the final performance.
	}
	
	\begin{table}[t]
		\centering
		\caption{Detection results with and w/o dense connections of CEM on COCO {\qcr{minival}} based on ResNet-50.}
		{
			\begin{tabular}{c|cccccc}
				\toprule
				Dense connections & AP & $\text{AP}_{50}$ & $\text{AP}_{75}$ & $\text{AP}_S$ & $\text{AP}_M$ & $\text{AP}_L$ \\
				\hline
				& 38.3  & 61.1  & 41.5  & 22.4  & 41.7  & 49.5 \\
				$\surd$  & 38.8  & 61.8  & 41.8  & 23.7  & 42.5  & 49.7 \\
				\bottomrule
			\end{tabular}%
		}
		\label{tab:dense_connection}%
	\end{table}%


	%

	\subsection{Experiments of Attention-guided Module}
	
	\subsubsection{Effectiveness of CxAM and CnAM}

	\begin{table}[t!]
		\centering
		\caption{Effects of our attention modules CxAM and CnAM with ResNet-50 on COCO {\qcr{minival}}.}
		\resizebox{1.0\linewidth}{!}{
			\begin{tabular}{c|cc|cccccc}
				\toprule
				& + CxAM & + CnAM & AP  & $\text{AP}_{50}$ & $\text{AP}_{75}$ & $\text{AP}_{S}$ & $\text{AP}_{M}$ & $\text{AP}_{L}$ \\
				\hline
				\multicolumn{1}{c|}{\multirow{4}[2]{*}{
						FPN}} &       &       & 39.3 & 61.9 & 42.6 & 23.6 & 42.9 & 50.4 \\
				& $\surd$     &       & 39.6 & 62.0 & 42.8 & 23.5 & 43.4 & 50.6 \\
				&       & $\surd$     & 39.8 & 62.4 & 43.2 & 23.8 & 43.3 & 51.6 \\
				& $\surd$    & $\surd$    & \textbf{40.1} & \textbf{62.5} & \textbf{43.2} & \textbf{23.9} & \textbf{43.6} & \textbf{52.4} \\
				\bottomrule
		\end{tabular}}
		\label{tab:effect_attention}%
	\end{table}%

	We conduct an ablation study to investigate the effects of our CxAM and CnAM. For FPN, we introduce our modules gradually and test the AP values of each combined models. 
    \myblue{As shown in Tab.~\ref{tab:effect_attention}, the model incorporated with}
	both CxAM and CnAM achieves the best performance in all \myblue{metrics}.  Moreover, if we only embed with one of them, the results also increase to some extent. These results demonstrate that our \myblue{modules} improve the performance \myblue{consistently} for the objects of all sizes by capturing much richer multi-scale information.

	
	\subsubsection{Visualization of Attention}
	
	\begin{figure*}[t]
		\centering
		{\scriptsize
			\resizebox{1.0\linewidth}{!}{
				\begin{tabular}{
						>{\centering\arraybackslash} m{10.5cm}
						>{\centering\arraybackslash} m{10.5cm}
						>{\centering\arraybackslash} m{10.5cm}
						>{\centering\arraybackslash} m{10.5cm}
						>{\centering\arraybackslash} m{10.5cm} 
						>{\centering\arraybackslash} m{10.5cm}}
					\includegraphics[height=70mm]{./000000063154.jpg}
					&\includegraphics[height=70mm]{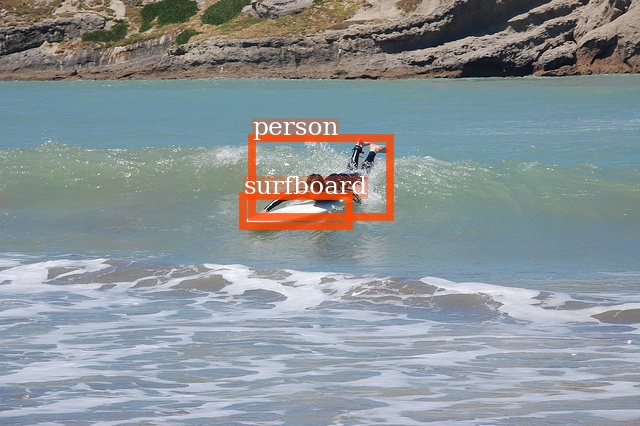}
					&\includegraphics[height=70mm]{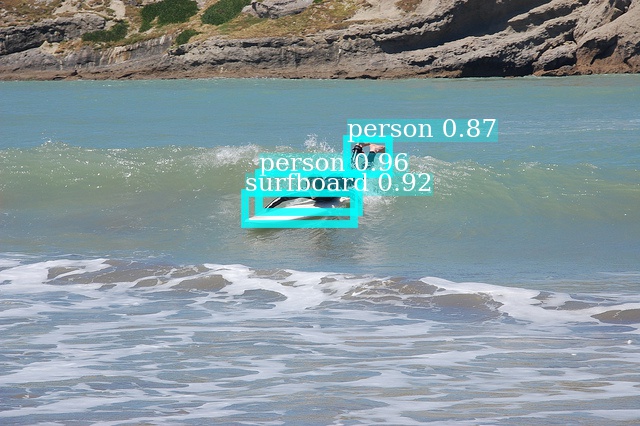}
					&\includegraphics[height=70mm]{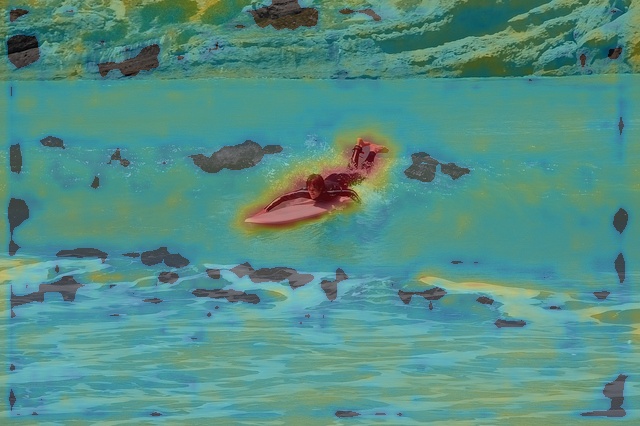}
					&\includegraphics[height=70mm]{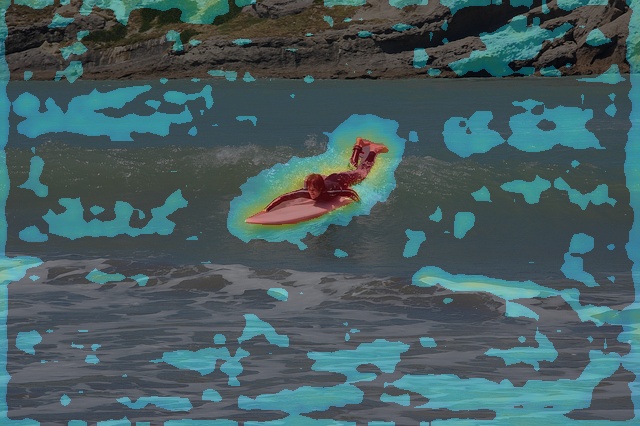}
					&\includegraphics[height=70mm]{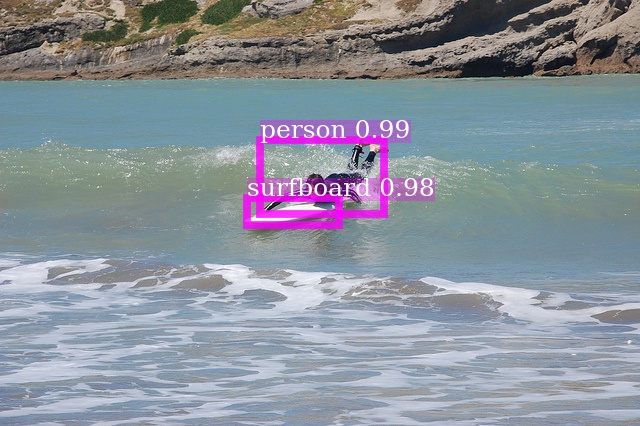}
					\\ \\
					\includegraphics[height=158mm]{./000000070158.jpg}
					&\includegraphics[height=158mm]{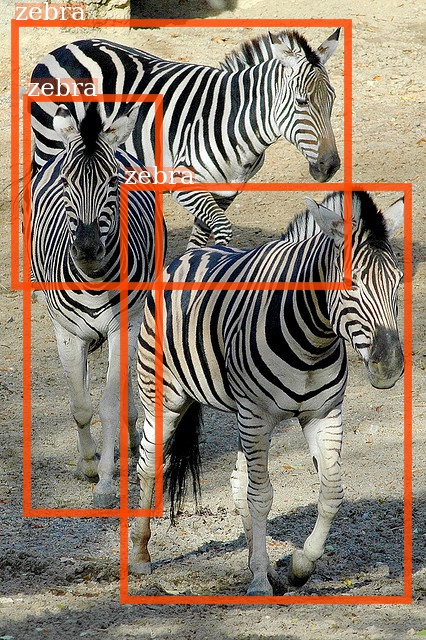}
					&\includegraphics[height=158mm]{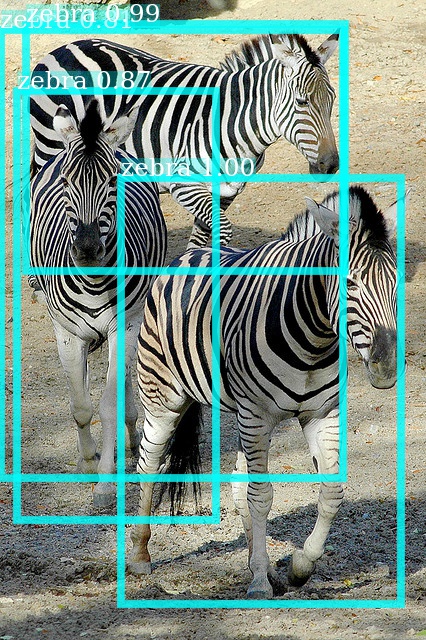}
					&\includegraphics[height=158mm]{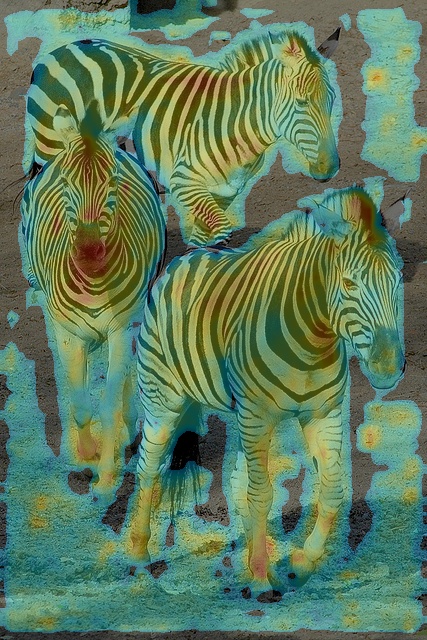}
					&\includegraphics[height=158mm]{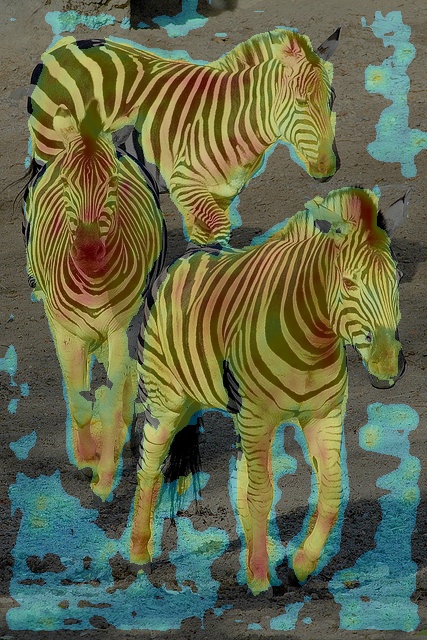}
					&\includegraphics[height=158mm]{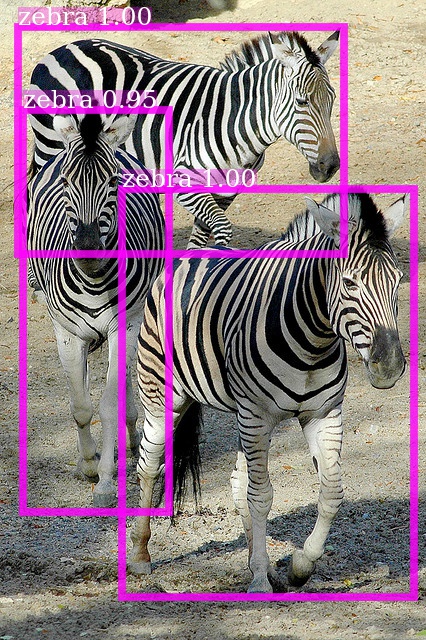}
					\\ \\
					\includegraphics[height=70mm]{./000000082715.jpg}
					&\includegraphics[height=70mm]{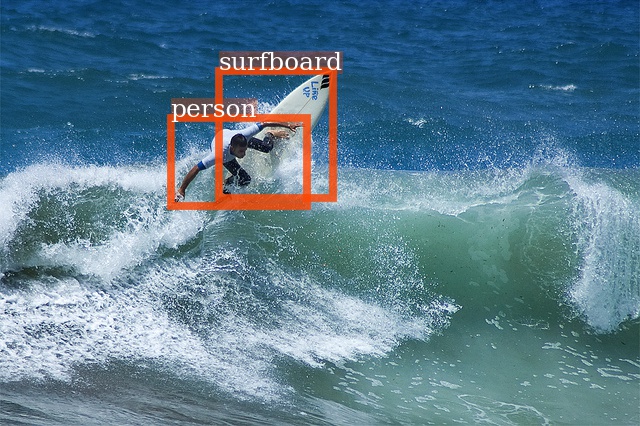}
					&\includegraphics[height=70mm]{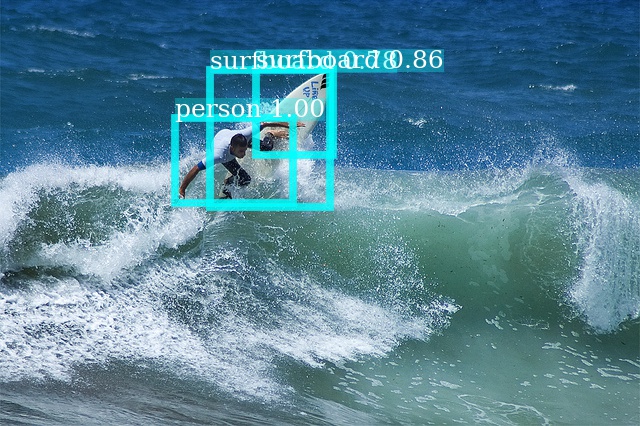}
					&\includegraphics[height=70mm]{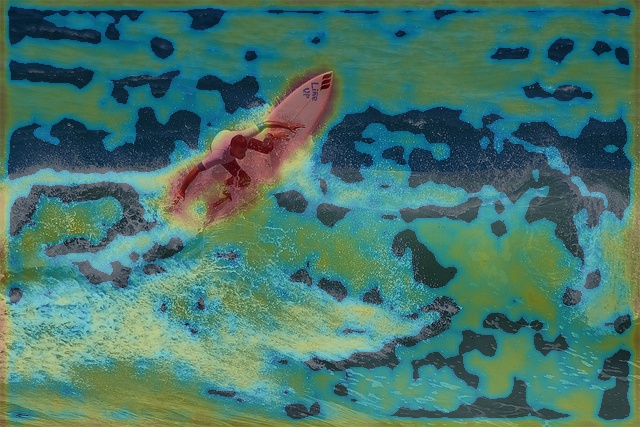}
					&\includegraphics[height=70mm]{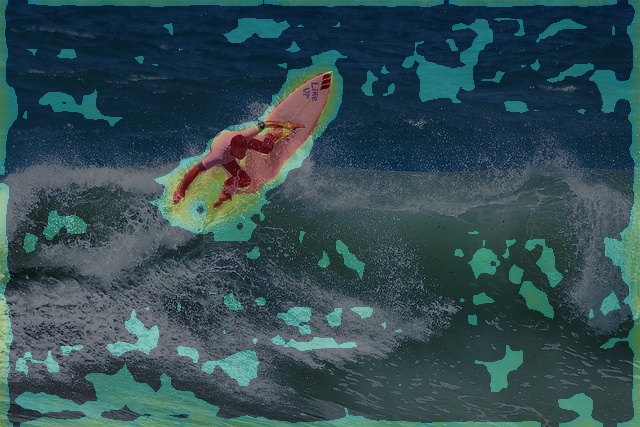}
					&\includegraphics[height=70mm]{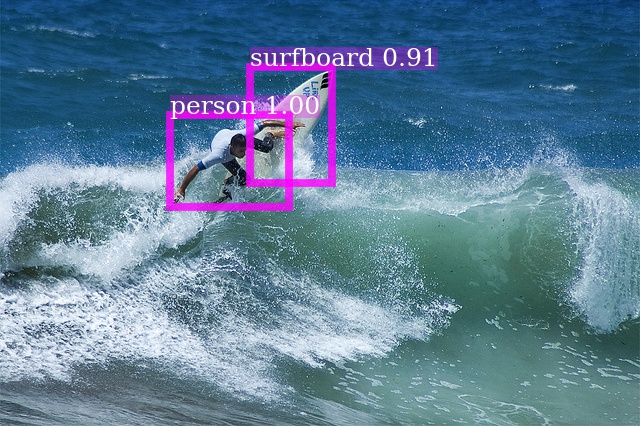}
					\\ \\
					\includegraphics[height=70mm]{./000000085823.jpg}
					&\includegraphics[height=70mm]{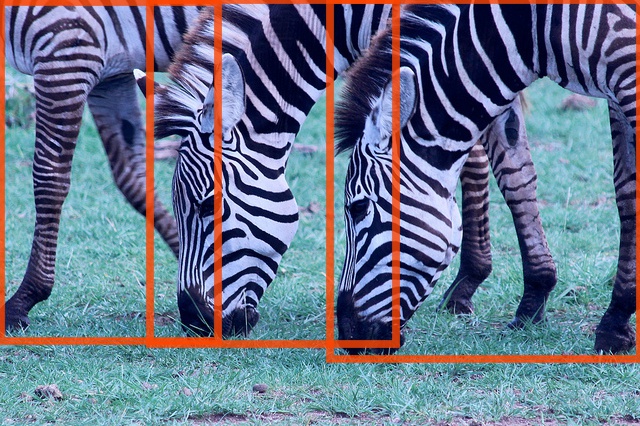}
					&\includegraphics[height=70mm]{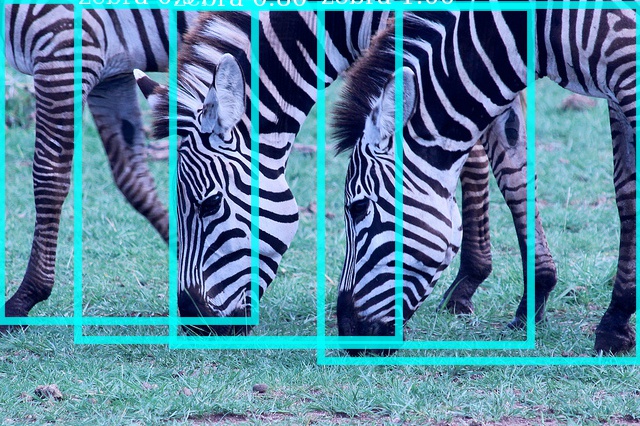}
					&\includegraphics[height=70mm]{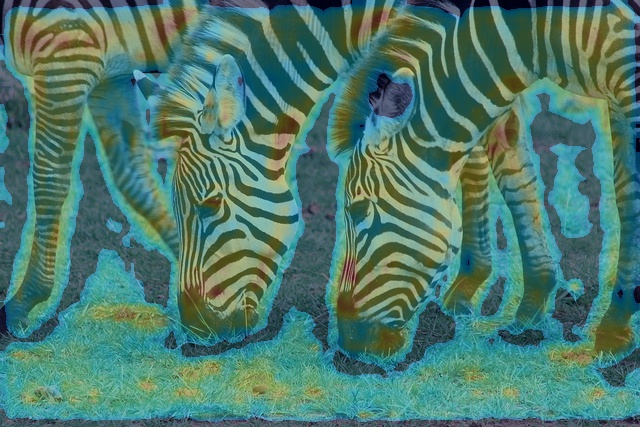}
					&\includegraphics[height=70mm]{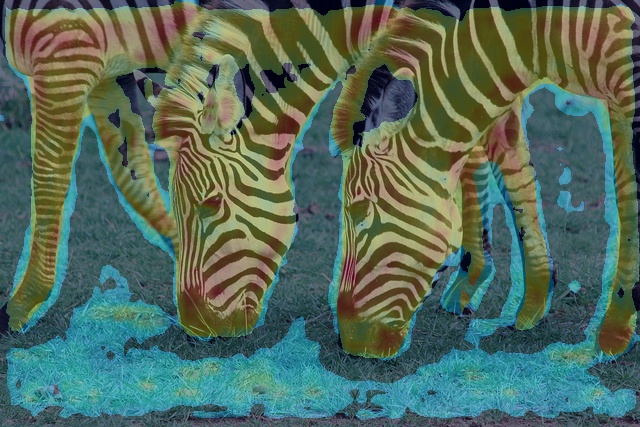}
					&\includegraphics[height=70mm]{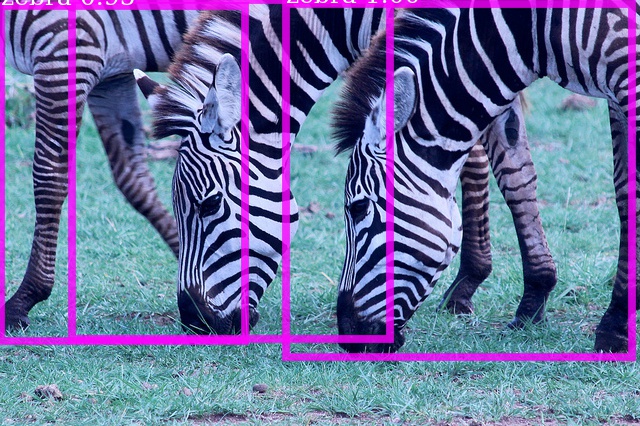}
					\\ \\
					\includegraphics[height=141mm]{./000000094326.jpg}
					&\includegraphics[height=141mm]{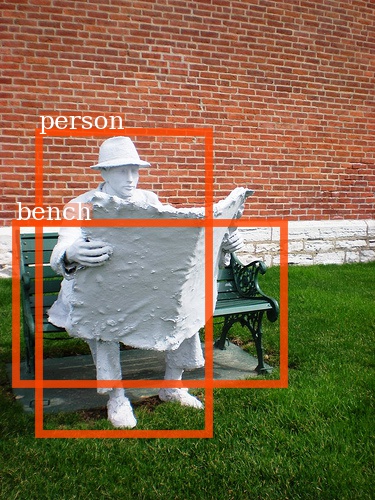}
					&\includegraphics[height=141mm]{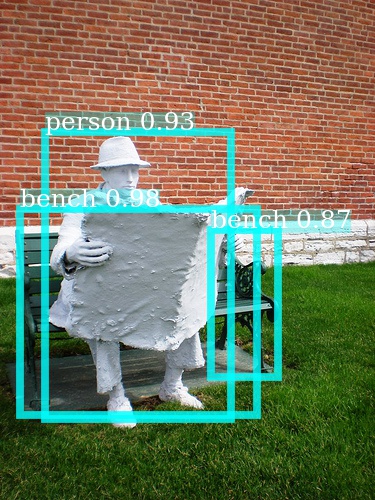}
					&\includegraphics[height=141mm]{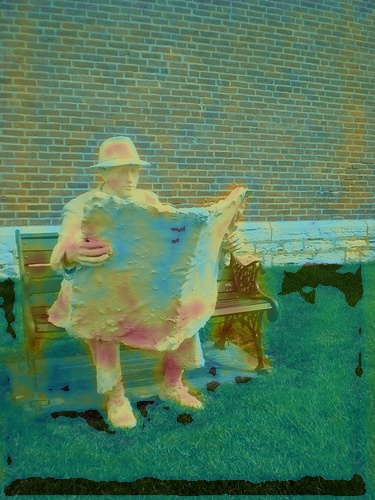}
					&\includegraphics[height=141mm]{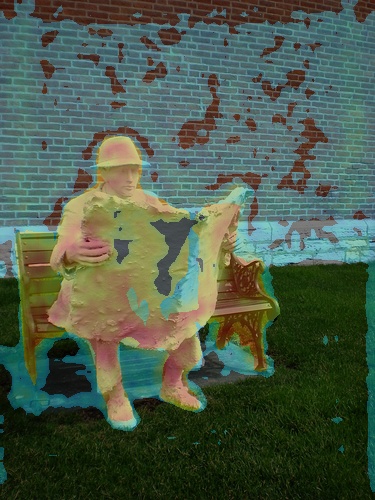}
					&\includegraphics[height=141mm]{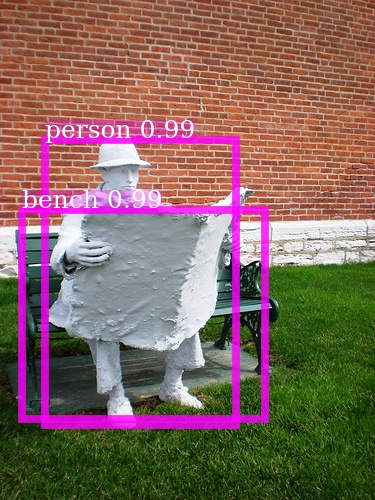}
					\\ \\
					\includegraphics[height=59mm]{./000000098716.jpg}
					&\includegraphics[height=59mm]{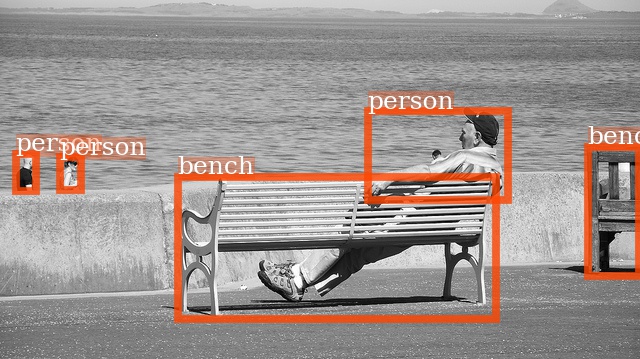}
					&\includegraphics[height=59mm]{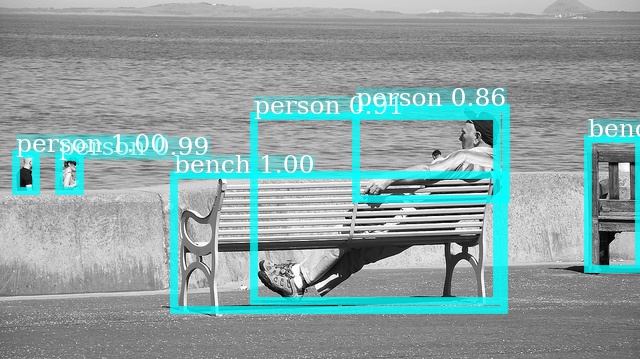}
					&\includegraphics[height=59mm]{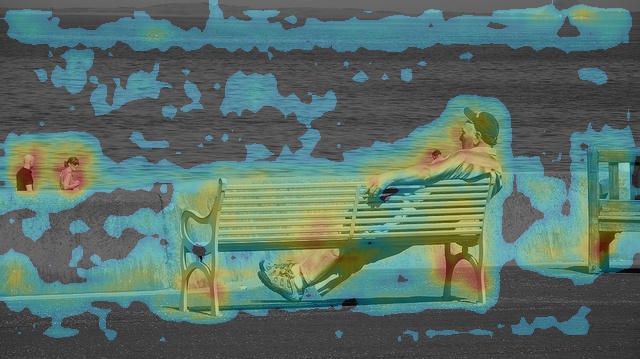}
					&\includegraphics[height=59mm]{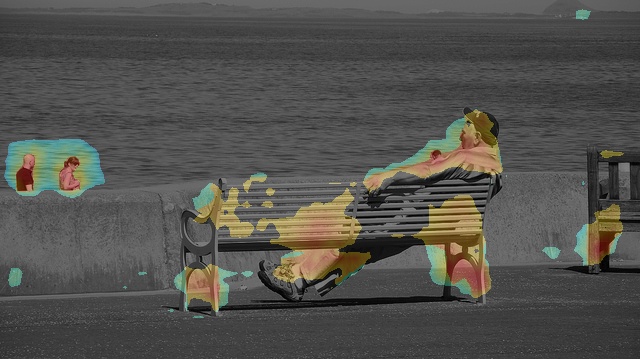}
					&\includegraphics[height=59mm]{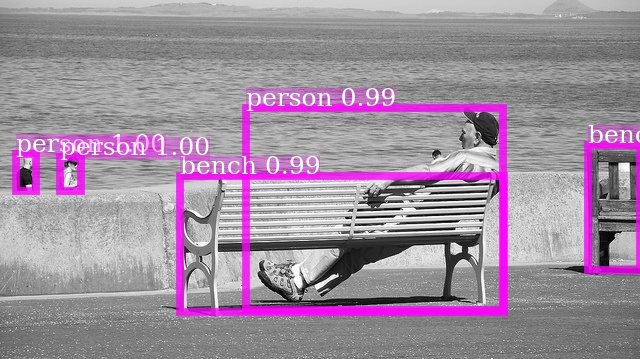}
					\\ \\
					\Huge{(a) Image} & \Huge{(b) Ground-truth} &\Huge{(c) CEM} & \Huge{(d) CEM + CnAM} & \Huge{(e) CEM + CnAM + CxAM} & \Huge{(f) Our result} 
			\end{tabular}}
		}
		\caption{Discussion for the impacts of our CxAM and CnAM modules via visualizing attention map on COCO {\qcr{minival}}.}
		\label{attention}
	\end{figure*}

	To further verify the effects of our attention modules, we pose the visual results of attention maps in Fig.~\ref{attention}.
	Compared to ground-truth bounding boxes (Fig.~\ref{attention} (b)), when only employing the CEM module (Fig.~\ref{attention} (c)), the results contain \myblue{some redundant} region proposals owing to the miscellaneous context information captured by CEM.
	After \myblue{incorporating} CnAM, as shown in Fig.~\ref{attention} (d), the attention model locates the \myblue{object much} more accurately. Nonetheless, the attention map also captures some unnecessary regions since the context information is insufficient to distinguish them. To handle this issue, we integrate the CxAM module, which contains stronger semantics, to further filter the needless dependencies and then obtain a clearer attention map (Fig.~\ref{attention} (e)). Therefore, in contrast to Fig.~\ref{attention} (c), our model (Fig.~\ref{attention} (f)) \myblue{gets rid of the redundant} regions and then achieves better performance. 

	\subsubsection{Impact of Pooling Method}
	
    \begin{table}[t]
      \centering
        \caption{Impact of pooling method in Attention-guided Module (AM).}
        \begin{tabular}{c|cc|c}
        \toprule
        \multirow{2}{*}{Backbone} & \multicolumn{2}{c|}{Attention-guided Module (AM)} & \multirow{2}{*}{AP} \\
        \cline{2-3}          & max pooling & avg pooling &  \\
        \hline
        \multirow{2}{*}{ResNet-50} & $\surd$     &       & 39.9 \\
              &       & $\surd$     & 40.1  \\
        \hline
        \multirow{2}{*}{ResNet-101} & $\surd$    &       & 41.6 \\
              &       & $\surd$     & 42.0 \\
        \bottomrule
        \end{tabular}%
      \label{tab:pooling}%
    \end{table}%
    
    Moreover, we also investigate the impact of pooling method in our Attention-guided Module (AM). From Tab.~\ref{tab:pooling}, the results show that compared with max pooling, average pooling obtains a similar but slightly better performance with both ResNet-50 and ResNet-101 backbones.

	\subsection{More Discussions}
	
	\subsubsection{Complexity Analysis}

	To analyze the model complexity, we compare the number of model parameters and 
    \myblue{floating-point operations per second (FLOPS).}
	First, we compare the number of \myblue{parameters} with or without the proposed modules (\ie, CEM and AM). 
	Tab.~\ref{tab:params} shows that AC-FCN improves the performance in a large margin while only introducing a few extra parameters.
	Besides, our proposed AC-FPN (ResNet-50), which contains fewer parameters than FPN (ResNet-101), can also obtain better performance. The results demonstrate that the improvement \myblue{brought by} our methods mainly comes from the elaborate design rather than the additional parameters.

	Furthermore, to evaluate the efficiency of our modules, we report the FLOPS of FPN and AC-FPN in Tab.~\ref{tab:flops}. Specifically, for feature maps $C_5$, we set dilation rate $= 1$ and stride $= 32$ in both FPN and AC-FPN.
	The results show that our modules improve performance significantly while increasing the computational cost slightly. Thus, we can draw a conclusion that the improvement mostly comes from our well-designed modules (\ie, CEM and AM) instead of the additional computations.

	In addition, we evaluate the training time of all the models using an NVIDIA Tesla P40 GPU.
    For Tab.~\ref{tab:params}, our ACFPN achieves more promising performance in both ResNet-50 and ResNet-101, and only requires $1.18$ seconds and $1.56$ seconds for each iteration, respectively.
    Moreover, we also investigate the change of training time when increasing the number of paths and dilation rate of ACFPN. As shown in Tab.~\ref{tab:dilation_rate}, even with more paths and dilation rates, the training cost grows very slightly.

	\subsubsection{Influence of Number of Path in CEM}

	To further evaluate the impacts of multi-scale context information, we adopt a different \myblue{number of paths} and dilation rates for our CEM module. From Tab.~\ref{tab:dilation_rate}, 
    \myblue{the model with too many (\eg, 7) or too few (\eg, 1) paths always achieves sub-optimal results.}
	This situation demonstrates that the features produced by fewer paths cannot sufficiently capture the information from different receptive fields very well. On the contrary, too many paths in CEM will cause the extracted features more complicated, which may contain much redundant information and thereby confuse the detector.
	Notably, compared to the results of $\text{path}=1$, using more paths can achieve better performance, \eg, the results of $\text{path}=3, 5, 7$. Thus, we can draw the conclusion that it is necessary and vital to increase the receptive field for larger input images. 
	%
	

	\begin{table}[t]
		\centering
		\caption{Detection results on COCO {\qcr{test-dev}} and the corresponding number of parameters and training time for each iteration.}
		\resizebox{0.95\linewidth}{!}
		{
			\begin{tabular}{c|ccc|ccc}
				\toprule
				\multirow{2}{*}{Backbone} & \multicolumn{3}{c|}{FPN} & \multicolumn{3}{c}{ACFPN} \\
				& \#Params & Time & AP    & \#Params & Time & AP \\
				\hline    ResNet-50 & 39.82M & 0.92s & 37.2  & 54.58M & 1.18s & 40.4 \\
				ResNet-101 & 57.94M & 1.24s & 39.4  & 72.69M & 1.56s & 42.4 \\
				\bottomrule
			\end{tabular}%
		}
		\label{tab:params}%
	\end{table}%
	
	\begin{table}[t]
		\centering
		\caption{FLOPS of detection results on COCO.}
		{
			\begin{tabular}{c|c|ccc}
				\toprule
				& Backbone & {\qcr{minival}} & {\qcr{test-dev}} & FLOPS \\
				\hline
				FPN   & \multirow{2}{*}{resnet50} & 36.7  & 37.2  & 97.63G \\
				ACFPN &       & 39.1  & 39.4  & 102.78G \\
				\hline
				FPN   & \multirow{2}{*}{resnet101} & 39.4  & 39.4  & 143.73G \\
				ACFPN &       & 41.5  & 41.7  & 148.89G \\
				\bottomrule
			\end{tabular}%
		}
		\label{tab:flops}%
	\end{table}%
	

	\begin{table}[t]
		\centering
		\caption{Influences of the path number and dilation rate in our CEM. We evaluate them with ResNet-50 on COCO {\qcr{minival}} and report the corresponding training time for each iteration.}
		\resizebox{1.0\linewidth}{!}
		{
			\begin{tabular}{c|c|cccccc|c}
				\toprule
				Path &  Rate & AP  & $\text{AP}_{50}$ & $\text{AP}_{75}$ & $\text{AP}_{S}$ & $\text{AP}_{M}$ & $\text{AP}_{L}$ & Time \\
				\hline
				%
				1 & (1) & 38.3 & 61.1 & 41.5 & 22.4 & 41.7 & 49.5 & 0.95s\\
				3 & (3,12,24) & 39.6  & 62.1  & 43.1  & 23.5  & 43.0  & 51.5 & 1.03s\\
				5 & (3,6,12,18,24) & \textbf{40.1}  & \textbf{62.5}  & 43.2  & \textbf{23.9}  & \textbf{43.6}  & \textbf{52.4} & 1.18s\\
				7 & (3,6,9,12,18,24,32) & 39.9  & 62.1  & \textbf{43.4}  & 23.6  & 43.1  & 52.0 & 1.38s\\
				\bottomrule
			\end{tabular}%
			\label{tab:dilation_rate}}
	\end{table}%

	\section{Experiments on Instance Segmentation}
	
	\begin{table*}[t]
		\centering
		\caption{Object detection
			results (bounding box AP) and Instance segmentation mask AP on COCO {\qcr{minival}}.}
		{
			\begin{tabular}{c|c|c|cccccc|cccccc}
				\toprule
				\multirow{2}[4]{*}{Methods} & \multirow{2}[4]{*}{Backbone} & \multicolumn{1}{r|}{\multirow{2}[4]{*}{ + Our modules}} & \multicolumn{6}{c|}{Object Detection}                & \multicolumn{6}{c}{Instance Sementation} \\
				\cline{4-15}   &       &       & AP    & $\text{AP}_{50}$ & $\text{AP}_{75}$ & $\text{AP}_{S}$ & $\text{AP}_{M}$ & $\text{AP}_{L}$ & AP   & $\text{AP}_{50}$ & $\text{AP}_{75}$ & $\text{AP}_{S}$ & $\text{AP}_{M}$ & $\text{AP}_{L}$ \\
				\hline
				\multirow{4}[4]{*}{Mask R-CNN} & \multirow{2}[2]{*}{ResNet-50} &       & 37.7  & 59.2  & 40.9  & 21.4  & 40.8  & 49.7  & 33.9  & 55.8  & 35.8  & 14.9  & 36.3  & 50.9 \\
				&       & $\surd$     &  40.7  &  62.8   & 44.0  &  24.2 &  44.1  &  53.0  &  36.0  &  59.4  &  38.1 & 17.0  &   39.2    & 53.2  \\
				\cline{2-15}          & \multirow{2}[2]{*}{ResNet-101} &       & 40.0 & 61.8  & 43.7 & 22.6 & 43.4 & 52.7 & 35.9 & 58.3  & 38.0 & 15.9 & 38.9 & 53.2 \\
				&       & $\surd$      & 42.8  & 65.2  & 47.0  & 26.2 & 46.6 & 54.2 & 38.0  & 61.9 & 40.1 & 18.0 & 41.5 &  54.6\\
				\hline
				\multirow{4}[4]{*}{PANet} & \multirow{2}[2]{*}{ResNet-50} &       & 39.5  & 60.4 &  42.6 & 23.9 & 43.3 & 49.6 &   35.2 &  57.4 & 37.2 & 16.6 & 38.6 & 50.8 \\
				&       & $\surd$      &  41.4 & 62.2 & 45.2  & 24.1 & 44.9 & 54.3  & 36.5 & 59.1   & 38.6 & 17.0 & 39.8  & 54.0 \\
				\cline{2-15}          & \multirow{2}[2]{*}{ResNet-101} &       & 41.5 & 62.1 & 45.1 & 24.1   & 45.2 & 54.4 & 36.7 & 59.1 & 38.8 & 16.5 & 40.1 & 54.9 \\
				&       & $\surd$      & 43.5 & 64.9 & 47.6 & 26.1 & 47.6 & 55.8 & 38.2 & 61.6 &     40.4  & 18.2 & 42.2 & 56.0  \\
				\hline
				\multirow{4}[4]{*}{Cascade R-CNN} & \multirow{2}[2]{*}{ResNet-50} &       & 41.3 & 59.6 & 44.9 & 23.1 & 44.2 & 55.4 & 35.4 & 56.2 & 37.8  & 15.7 & 37.6 & 53.4 \\
				&       & $\surd$      & 43.6 & 62.7 & 47.8 & 25.4 & 47.0 & 57.5 & 37.2 & 59.5 & 39.7 &   17.4 & 40.2 & 54.7 \\
				\cline{2-15}          & \multirow{2}[2]{*}{ResNet-101} &       & 43.3 & 61.7 & 47.3 & 24.2 & 46.3 & 58.2 & 37.1  & 58.6 & 39.8 & 16.7 & 39.7 & 55.7 \\
				&       & $\surd$      &  45.4 & 64.5 & 49.3 & 27.6 & 49.1 & 58.6 & 38.5 & 61.2 &    41.0 & 18.3 & 41.8 & 56.1 \\
				\bottomrule
		\end{tabular}}
		\label{tab:instance_segmentation_minival}%
	\end{table*}%

	\begin{table*}[t]
		\centering
		\footnotesize
		\caption{Object detection 
			results (bounding box AP) and instance segmentation mask AP on COCO {\qcr{test-dev}}.}
		{
			\begin{tabular}{c|c|c|cccccc|cccccc}
				\toprule
				\multirow{2}{*}{Methods} & \multirow{2}{*}{Backbone} & \multicolumn{1}{c|}{\multirow{2}{*}{ + Our modules}} & \multicolumn{6}{c|}{Object Detection}                & \multicolumn{6}{c}{Instance Segmentation} \\
				\cline{4-15}   &       &       & AP    & $\text{AP}_{50}$ & $\text{AP}_{75}$ & $\text{AP}_{S}$ & $\text{AP}_{M}$ & $\text{AP}_{L}$ & AP   & $\text{AP}_{50}$ & $\text{AP}_{75}$ & $\text{AP}_{S}$ & $\text{AP}_{M}$ & $\text{AP}_{L}$ \\
				\hline
				\multirow{4}{*}{Mask R-CNN} & \multirow{2}[2]{*}{ResNet-50} &       &  38.0  & 59.7 & 41.3 & 21.2  & 40.2  & 48.1  & 34.2  & 56.4 & 36.0  &   14.8 & 36.0 & 49.7 \\
				&       & $\surd$     & 41.2 & 63.5  & 44.9 & 23.8 & 43.6 & 52.2 & 36.4 & 59.9 & 38.6 &  16.7 & 38.8 &  52.5\\
				\cline{2-15}          & \multirow{2}[2]{*}{ResNet-101} &       & 40.2 & 62.0 & 43.9 & 22.8   & 43.0 & 51.1  &35.9 & 58.5 & 38.0 & 15.9 & 38.2 & 51.8 \\
				&       & $\surd$      & 43.1 & 65.5 & 47.2 & 25.5  & 45.9 & 54.3 & 38.3 & 62.0 & 40.7 & 18.1 & 41.0 & 54.6 \\
				\hline
				\multirow{4}{*}{PANet} & \multirow{2}[2]{*}{ResNet-50} &       & 40.0 & 60.7 &  43.6 & 22.6 & 42.7 & 50.3 & 35.5 & 57.6 & 37.6 & 15.6  & 37.9 & 51.3 \\
				&   & $\surd$      & 41.8 & 62.9 & 45.8 & 24.0  & 44.4 & 52.7 & 36.9 & 59.8 & 39.2 &  16.8 & 392 & 53.3 \\
				\cline{2-15}          & \multirow{2}[2]{*}{ResNet-101} &       & 41.8  & 62.7 & 45.7 & 23.6 & 44.7 & 52.7 & 37.0 & 59.7 & 39.3 & 16.5 & 39.7  & 53.4\\
				&       & $\surd$ & 43.6 & 64.7 & 47.7 & 25.7 & 46.8 & 54.5 & 38.3 & 61.7 & 40.7 & 18.0 & 41.1 & 54.7\\
				\hline
				\multirow{4}{*}{Cascade R-CNN} & \multirow{2}[2]{*}{ResNet-50} &       &  41.7  & 60.0 & 45.4 &  23.1 &  43.6 & 54.2 & 35.6 & 57.0 & 38.0 & 15.5 & 37.1 & 52.0 \\
				&       & $\surd$      & 43.8 & 62.9 & 47.8 & 25.2 & 46.2 & 55.9 & 37.4 & 59.9 & 40.0 &  17.2  & 39.6 &  53.6\\
				\cline{2-15}          & \multirow{2}[2]{*}{ResNet-101} &       & 43.4  & 61.9 & 47.2 &  23.9 & 45.9 & 56.2 & 37.1 & 58.9 & 39.7 & 16.2 & 39.1 & 54.1 \\
				&       & $\surd$      & 45.6 & 64.9 & 49.8 & 27.2 & 48.6 & 57.5 & 38.8 & 61.7 & 41.7 & 18.8 & 41.4 & 55.1 \\
				\bottomrule
		\end{tabular}}
		\label{tab:instance_segmentation_test_dev}%
	\end{table*}%

	To further verify the \myblue{adaptability} of our CEM and AM, we extend them to some instance segmentation models, \qi{including} Mask R-CNN~\cite{he2017mask}, PANet~\cite{liu2018path} and Cascade R-CNN~\cite{cai2018cascade}.
	Specifically, we simply extend the Cascade R-CNN model to instance segmentation by adding the mask branch following~\cite{he2017mask}. 
	Moreover, we use the official implementations and evaluate the models with the aforementioned metrics, including AP, $\text{AP}_{50}$, $\text{AP}_{75}$, $\text{AP}_{S}$, $\text{AP}_{M}$ and $\text{AP}_{L}$.

	\subsection{Instance Segmentation Results}
	
	We investigate the performance of CEM and AM in terms of instance segmentation by plugging them into existing segmentation networks. 
	\qi{Then we test these incremental models on {\qcr{minival}} and {\qcr{test-dev}}, and report the results in Tabs.~\ref{tab:instance_segmentation_minival} and~\ref{tab:instance_segmentation_test_dev}, respectively.}
	
	As shown in Tabs.~\ref{tab:instance_segmentation_minival} and~\ref{tab:instance_segmentation_test_dev}, the models with \myblue{the proposed CEM and AM} achieve more promising performance compared to the original ones. To be specific, due to the lack of rich context information, 
	Cascade R-CNN gets a limited performance in detection and thereby obtains the inferior results in segmentation.
	Thus, the rich context information from various receptive fields is critical for both object detection and instance segmentation tasks.
	
	\subsection{Object Detection Results}
	
	To further validate the effects of our modules, we report the intermediate results of instance segmentation, \ie, APs of object detection. 
	\myblue{As} shown in Tabs.~\ref{tab:instance_segmentation_minival} and~\ref{tab:instance_segmentation_test_dev}, 
    \myblue{the models integrated with our modules achieve much better performance in all metrics, which demonstrates}
	that our proposed modules capture more \myblue{effective} information 
    \myblue{and thereby greatly facilitate the}
	bounding box regression and object recognition.

	\subsection{Visualization Results}

	\begin{figure*}[t]
		\centering
		{\scriptsize
			\resizebox{1.0\linewidth}{!}{
				\begin{tabular}{>{\centering\arraybackslash} m{5.5cm}
						>{\centering\arraybackslash} m{7.9cm}
						>{\centering\arraybackslash} m{7.9cm}
						>{\centering\arraybackslash} m{8.95cm} >{\centering\arraybackslash} m{4.4cm}
						>{\centering\arraybackslash} m{8cm}}
					\huge{Mask R-CNN w/o CEM and AM}
					&\includegraphics[height=60mm]{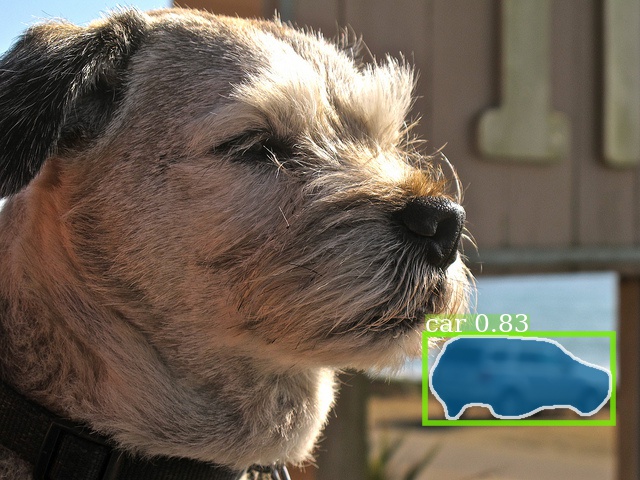} &\includegraphics[height=60mm]{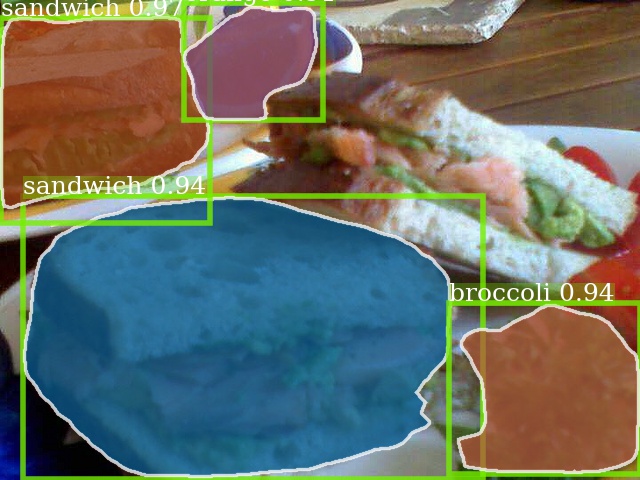}
					&\includegraphics[height=60mm]{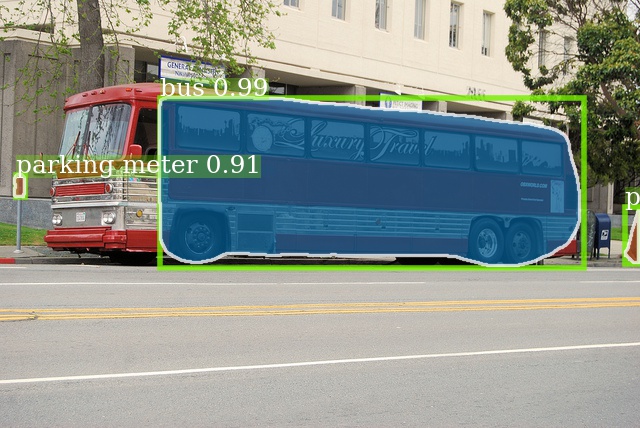}
					&\includegraphics[height=60mm]{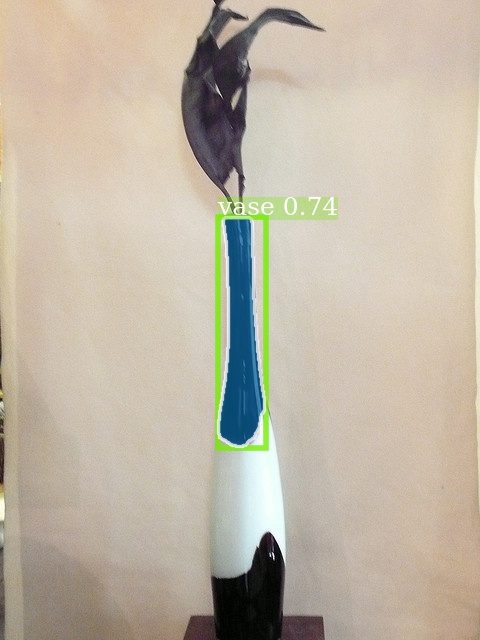}
					&\includegraphics[height=60mm]{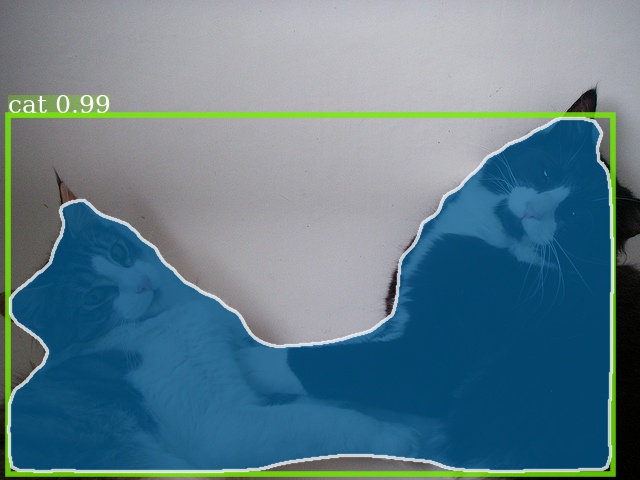}
					\\ \\
					\huge{Mask R-CNN w/ CEM and AM}
					&\includegraphics[height=60mm]{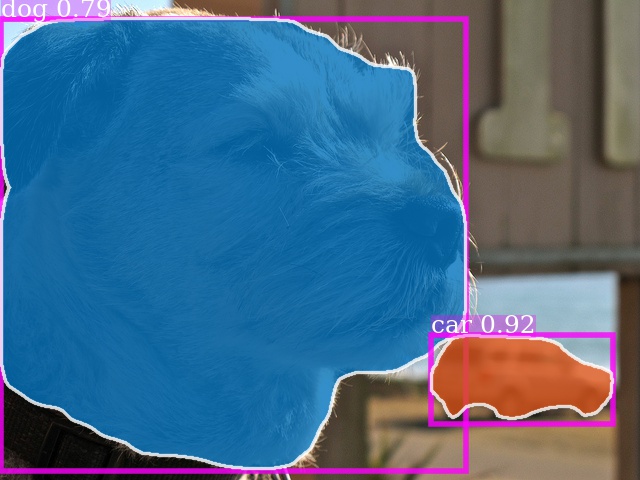} &\includegraphics[height=60mm]{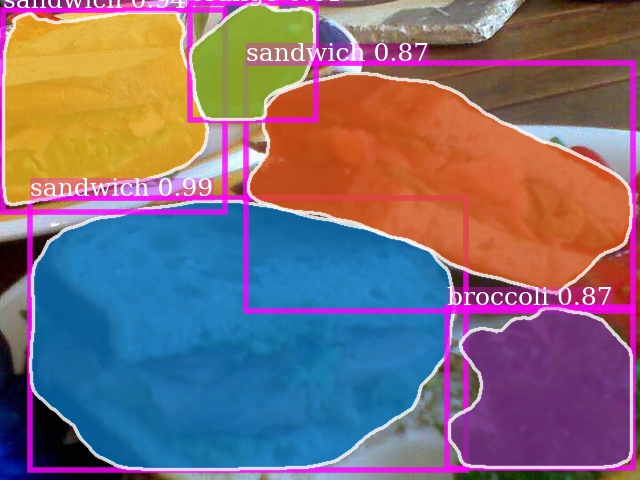}
					&\includegraphics[height=60mm]{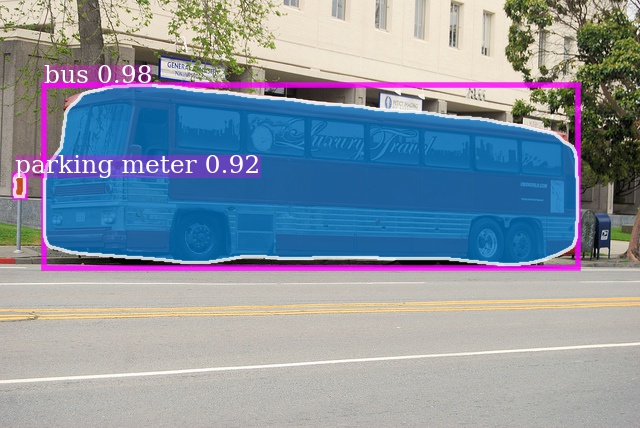}
					&\includegraphics[height=60mm]{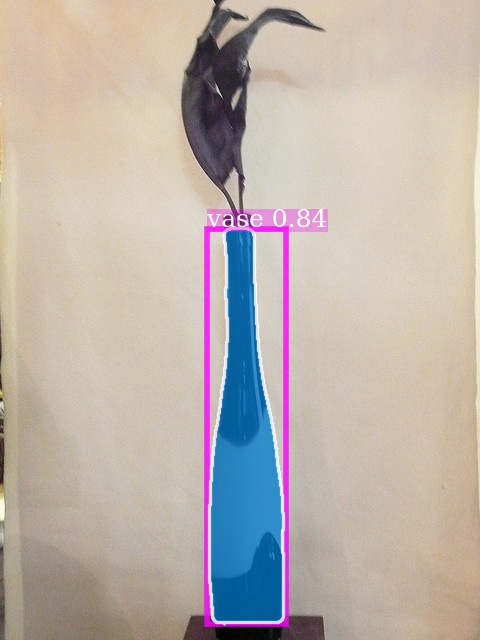}
					&\includegraphics[height=60mm]{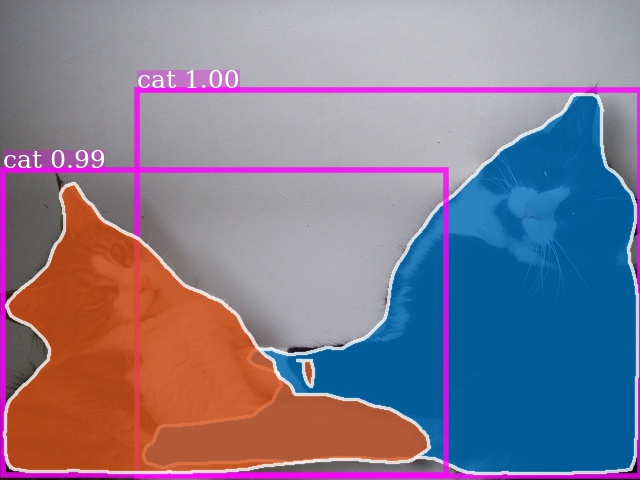}
					\\ \\
					& \Huge{(a)} & \Huge{(b)} & \Huge{(c)} & \Huge{(d)} & \Huge{(e)}
			\end{tabular}}
		}
		\caption{Results of Mask R-CNN with (w) and without (w/o) our modules built upon ResNet-50 on COCO {\qcr{minival}}.}
		\label{instance_segmentation_visualization}
	\end{figure*}
	
	We further show the visual results of Mask R-CNN with or without our CEM and AM. For a fair comparison, we build the models upon ResNet-50 and test 
	on COCO {\qcr{minival}}.
	The results with $\text{threshold}=0.7$ are exhibited in Fig.~\ref{instance_segmentation_visualization}.
	Fig.~\ref{instance_segmentation_visualization} (a) and (b) show that the model with CEM and AM is able to capture the large objects, which are \myblue{always} neglected by the original Mask R-CNN. Moreover, \myblue{as} shown in Fig.~\ref{instance_segmentation_visualization} (c) and (d), although Mask R-CNN \myblue{successfully detects the} large objects, \myblue{the bounding boxes are always imprecise due to the limitation of receptive fields.}
	Furthermore, from Fig.~\ref{instance_segmentation_visualization} (e), based on the more discriminative features, Mask R-CNN with our modules distinguishes the instances accurately while the counterpart \myblue{fails}.

	
	\section{Conclusion}
	
	
	In this paper, we build a novel architecture, \myblue{named} AC-FPN, containing two main sub-modules (CEM and AM), to solve the contradictory requirements between feature map resolution and receptive fields in high-resolution images, and enhances the discriminative ability of feature representations.
	Moreover, our proposed modules (\ie, CEM and AM) can be readily plugged into the existing object detection and segmentation networks, and be easily trained end-to-end.
	The extensive experiments on object detection and instance segmentation tasks demonstrate the superiority and adaptability of our CEM and AM modules.

	\ifCLASSOPTIONcaptionsoff
	\newpage
	\fi


\begin{thebibliography}{10}
	
	\bibitem{jiang2017face}
	Huaizu Jiang and Erik Learned-Miller.
	\newblock Face detection with the faster r-cnn.
	\newblock In {\em IEEE International Conference on Automatic Face \& Gesture
		Recognition}, pages 650--657. IEEE, 2017.
	
	\bibitem{yang2016wider}
	Shuo Yang, Ping Luo, Chen-Change Loy, and Xiaoou Tang.
	\newblock Wider face: A face detection benchmark.
	\newblock In {\em IEEE Conference on Computer Vision and Pattern Recognition},
	pages 5525--5533, 2016.
	
	\bibitem{yang2018faceness}
	Shuo Yang, Ping Luo, Chen~Change Loy, and Xiaoou Tang.
	\newblock Faceness-net: Face detection through deep facial part responses.
	\newblock {\em IEEE Transactions on Pattern Analysis and Machine Intelligence},
	40(8):1845--1859, 2018.
	
	\bibitem{shami2018people}
	Mamoona Shami, Salman Maqbool, Hasan Sajid, Yasar Ayaz, and Sen-Ching~Samson
	Cheung.
	\newblock People counting in dense crowd images using sparse head detections.
	\newblock {\em IEEE Transactions on Circuits and Systems for Video Technology},
	2018.
	
	\bibitem{stewart2016end}
	Russell Stewart, Mykhaylo Andriluka, and Andrew~Y Ng.
	\newblock End-to-end people detection in crowded scenes.
	\newblock In {\em IEEE Conference on Computer Vision and Pattern Recognition},
	pages 2325--2333, 2016.
	
	\bibitem{wang2018pedestrian}
	Shiguang Wang, Jian Cheng, Haijun Liu, Feng Wang, and Hui Zhou.
	\newblock Pedestrian detection via body part semantic and contextual
	information with dnn.
	\newblock {\em IEEE Transactions on Multimedia}, 20(11):3148--3159, 2018.
	
	\bibitem{li2017scale}
	Jianan Li, Xiaodan Liang, ShengMei Shen, Tingfa Xu, Jiashi Feng, and Shuicheng
	Yan.
	\newblock Scale-aware fast r-cnn for pedestrian detection.
	\newblock {\em IEEE Transactions on Multimedia}, 20(4):985--996, 2017.
	
	\bibitem{qian2019oriented}
	Yeqiang Qian, Ming Yang, Xu~Zhao, Chunxiang Wang, and Bing Wang.
	\newblock Oriented spatial transformer network for pedestrian detection using
	fish-eye camera.
	\newblock {\em IEEE Transactions on Multimedia}, 2019.
	
	\bibitem{feichtenhofer2017detect}
	Christoph Feichtenhofer, Axel Pinz, and Andrew Zisserman.
	\newblock Detect to track and track to detect.
	\newblock In {\em IEEE International Conference on Computer Vision}, pages
	3038--3046, 2017.
	
	\bibitem{yuan2017incremental}
	Yuan Yuan, Zhitong Xiong, and Qi~Wang.
	\newblock An incremental framework for video-based traffic sign detection,
	tracking, and recognition.
	\newblock {\em IEEE Transactions on Intelligent Transportation Systems},
	18(7):1918--1929, 2017.
	
	\bibitem{chen2018learning}
	Kai Chen and Wenbing Tao.
	\newblock Learning linear regression via single-convolutional layer for visual
	object tracking.
	\newblock {\em IEEE Transactions on Multimedia}, 21(1):86--97, 2018.
	
	\bibitem{hu2018robust}
	Hongwei Hu, Bo~Ma, Jianbing Shen, Hanqiu Sun, Ling Shao, and Fatih Porikli.
	\newblock Robust object tracking using manifold regularized convolutional
	neural networks.
	\newblock {\em IEEE Transactions on Multimedia}, 21(2):510--521, 2018.
	
	\bibitem{wang2018learning}
	Qiurui Wang, Chun Yuan, Jingdong Wang, and Wenjun Zeng.
	\newblock Learning attentional recurrent neural network for visual tracking.
	\newblock {\em IEEE Transactions on Multimedia}, 21(4):930--942, 2018.
	
	\bibitem{ren2015faster}
	Shaoqing Ren, Kaiming He, Ross Girshick, and Jian Sun.
	\newblock Faster r-cnn: Towards real-time object detection with region proposal
	networks.
	\newblock In {\em Advances in Neural Information Processing Systems}, pages
	91--99, 2015.
	
	\bibitem{lin2017focal}
	Tsung-Yi Lin, Priya Goyal, Ross Girshick, Kaiming He, and Piotr Doll{\'a}r.
	\newblock Focal loss for dense object detection.
	\newblock In {\em IEEE International Conference on Computer Vision}, pages
	2980--2988, 2017.
	
	\bibitem{li2018detnet}
	Zeming Li, Chao Peng, Gang Yu, Xiangyu Zhang, Yangdong Deng, and Jian Sun.
	\newblock Detnet: Design backbone for object detection.
	\newblock In {\em European Conference on Computer Vision}, pages 334--350,
	2018.
	
	\bibitem{he2016deep}
	Kaiming He, Xiangyu Zhang, Shaoqing Ren, and Jian Sun.
	\newblock Deep residual learning for image recognition.
	\newblock In {\em IEEE Conference on Computer Vision and Pattern Recognition},
	pages 770--778, 2016.
	
	\bibitem{lin2017feature}
	Tsung-Yi Lin, Piotr Doll{\'a}r, Ross Girshick, Kaiming He, Bharath Hariharan,
	and Serge Belongie.
	\newblock Feature pyramid networks for object detection.
	\newblock In {\em IEEE Conference on Computer Vision and Pattern Recognition},
	pages 2117--2125, 2017.
	
	\bibitem{liu2018path}
	Shu Liu, Lu~Qi, Haifang Qin, Jianping Shi, and Jiaya Jia.
	\newblock Path aggregation network for instance segmentation.
	\newblock In {\em IEEE Conference on Computer Vision and Pattern Recognition},
	pages 8759--8768, 2018.
	
	\bibitem{girshick2014rich}
	Ross Girshick, Jeff Donahue, Trevor Darrell, and Jitendra Malik.
	\newblock Rich feature hierarchies for accurate object detection and semantic
	segmentation.
	\newblock In {\em IEEE Conference on Computer Vision and Pattern Recognition},
	pages 580--587, 2014.
	
	\bibitem{uijlings2013selective}
	Jasper~RR Uijlings, Koen~EA Van De~Sande, Theo Gevers, and Arnold~WM Smeulders.
	\newblock Selective search for object recognition.
	\newblock {\em International Journal of Computer Vision}, 104(2):154--171,
	2013.
	
	\bibitem{girshick2015fast}
	Ross Girshick.
	\newblock Fast r-cnn.
	\newblock In {\em IEEE International Conference on Computer Vision}, pages
	1440--1448, 2015.
	
	\bibitem{dai2016r}
	Jifeng Dai, Yi~Li, Kaiming He, and Jian Sun.
	\newblock R-fcn: Object detection via region-based fully convolutional
	networks.
	\newblock In {\em Advances in Neural Information Processing Systems}, pages
	379--387, 2016.
	
	\bibitem{he2017mask}
	Kaiming He, Georgia Gkioxari, Piotr Doll{\'a}r, and Ross Girshick.
	\newblock Mask r-cnn.
	\newblock In {\em IEEE International Conference on Computer Vision}, pages
	2961--2969, 2017.
	
	\bibitem{cai2018cascade}
	Zhaowei Cai and Nuno Vasconcelos.
	\newblock Cascade r-cnn: Delving into high quality object detection.
	\newblock In {\em IEEE Conference on Computer Vision and Pattern Recognition},
	pages 6154--6162, 2018.
	
	\bibitem{liu2019cbnet}
	Yudong Liu, Yongtao Wang, Siwei Wang, TingTing Liang, Qijie Zhao, Zhi Tang, and
	Haibin Ling.
	\newblock Cbnet: A novel composite backbone network architecture for object
	detection.
	\newblock {\em arXiv preprint arXiv:1909.03625}, 2019.
	
	\bibitem{sermanet2013overfeat}
	Pierre Sermanet, David Eigen, Xiang Zhang, Micha{\"e}l Mathieu, Rob Fergus, and
	Yann LeCun.
	\newblock Overfeat: Integrated recognition, localization and detection using
	convolutional networks.
	\newblock {\em arXiv preprint arXiv:1312.6229}, 2013.
	
	\bibitem{redmon2016you}
	Joseph Redmon, Santosh Divvala, Ross Girshick, and Ali Farhadi.
	\newblock You only look once: Unified, real-time object detection.
	\newblock In {\em IEEE Conference on Computer Vision and Pattern Recognition},
	pages 779--788, 2016.
	
	\bibitem{liu2016ssd}
	Wei Liu, Dragomir Anguelov, Dumitru Erhan, Christian Szegedy, Scott Reed,
	Cheng-Yang Fu, and Alexander~C Berg.
	\newblock Ssd: Single shot multibox detector.
	\newblock In {\em European Conference on Computer Vision}, pages 21--37.
	Springer, 2016.
	
	\bibitem{redmon2017yolo9000}
	Joseph Redmon and Ali Farhadi.
	\newblock Yolo9000: better, faster, stronger.
	\newblock In {\em IEEE Conference on Computer Vision and Pattern Recognition},
	pages 7263--7271, 2017.
	
	\bibitem{fu2017dssd}
	Cheng-Yang Fu, Wei Liu, Ananth Ranga, Ambrish Tyagi, and Alexander~C Berg.
	\newblock Dssd: Deconvolutional single shot detector.
	\newblock {\em arXiv preprint arXiv:1701.06659}, 2017.
	
	\bibitem{shen2017dsod}
	Zhiqiang Shen, Zhuang Liu, Jianguo Li, Yu-Gang Jiang, Yurong Chen, and
	Xiangyang Xue.
	\newblock Dsod: Learning deeply supervised object detectors from scratch.
	\newblock In {\em IEEE International Conference on Computer Vision}, pages
	1919--1927, 2017.
	
	\bibitem{bell2016inside}
	Sean Bell, C~Lawrence~Zitnick, Kavita Bala, and Ross Girshick.
	\newblock Inside-outside net: Detecting objects in context with skip pooling
	and recurrent neural networks.
	\newblock In {\em IEEE Conference on Computer Vision and Pattern Recognition},
	pages 2874--2883, 2016.
	
	\bibitem{chen2018context}
	Zhe Chen, Shaoli Huang, and Dacheng Tao.
	\newblock Context refinement for object detection.
	\newblock In {\em European Conference on Computer Vision}, pages 71--86, 2018.
	
	\bibitem{hu2018relation}
	Han Hu, Jiayuan Gu, Zheng Zhang, Jifeng Dai, and Yichen Wei.
	\newblock Relation networks for object detection.
	\newblock In {\em IEEE Conference on Computer Vision and Pattern Recognition},
	pages 3588--3597, 2018.
	
	\bibitem{chen2017spatial}
	Xinlei Chen and Abhinav Gupta.
	\newblock Spatial memory for context reasoning in object detection.
	\newblock In {\em IEEE International Conference on Computer Vision}, pages
	4086--4096, 2017.
	
	\bibitem{li2017attentive}
	Jianan Li, Yunchao Wei, Xiaodan Liang, Jian Dong, Tingfa Xu, Jiashi Feng, and
	Shuicheng Yan.
	\newblock Attentive contexts for object detection.
	\newblock {\em IEEE Transactions on Multimedia}, 19(5):944--954, 2017.
	
	\bibitem{chen2018iterative}
	Xinlei Chen, Li-Jia Li, Li~Fei-Fei, and Abhinav Gupta.
	\newblock Iterative visual reasoning beyond convolutions.
	\newblock In {\em IEEE Conference on Computer Vision and Pattern Recognition},
	pages 7239--7248, 2018.
	
	\bibitem{mnih2014recurrent}
	Volodymyr Mnih, Nicolas Heess, Alex Graves, et~al.
	\newblock Recurrent models of visual attention.
	\newblock In {\em Advances in Neural Information Processing Systems}, pages
	2204--2212, 2014.
	
	\bibitem{wang2017residual}
	Fei Wang, Mengqing Jiang, Chen Qian, Shuo Yang, Cheng Li, Honggang Zhang,
	Xiaogang Wang, and Xiaoou Tang.
	\newblock Residual attention network for image classification.
	\newblock In {\em IEEE Conference on Computer Vision and Pattern Recognition},
	pages 3156--3164, 2017.
	
	\bibitem{chen2016attention}
	Liang-Chieh Chen, Yi~Yang, Jiang Wang, Wei Xu, and Alan~L Yuille.
	\newblock Attention to scale: Scale-aware semantic image segmentation.
	\newblock In {\em IEEE Conference on Computer Vision and Pattern Recognition},
	pages 3640--3649, 2016.
	
	\bibitem{ren2017end}
	Mengye Ren and Richard~S Zemel.
	\newblock End-to-end instance segmentation with recurrent attention.
	\newblock In {\em IEEE Conference on Computer Vision and Pattern Recognition},
	pages 6656--6664, 2017.
	
	\bibitem{lu2017knowing}
	Jiasen Lu, Caiming Xiong, Devi Parikh, and Richard Socher.
	\newblock Knowing when to look: Adaptive attention via a visual sentinel for
	image captioning.
	\newblock In {\em IEEE Conference on Computer Vision and Pattern Recognition},
	pages 375--383, 2017.
	
	\bibitem{xu2015show}
	Kelvin Xu, Jimmy Ba, Ryan Kiros, Kyunghyun Cho, Aaron Courville, Ruslan
	Salakhudinov, Rich Zemel, and Yoshua Bengio.
	\newblock Show, attend and tell: Neural image caption generation with visual
	attention.
	\newblock In {\em International Conference on Machine Learning}, pages
	2048--2057, 2015.
	
	\bibitem{you2016image}
	Quanzeng You, Hailin Jin, Zhaowen Wang, Chen Fang, and Jiebo Luo.
	\newblock Image captioning with semantic attention.
	\newblock In {\em IEEE Conference on Computer Vision and Pattern Recognition},
	pages 4651--4659, 2016.
	
	\bibitem{bahdanau2014neural}
	Dzmitry Bahdanau, Kyunghyun Cho, and Yoshua Bengio.
	\newblock Neural machine translation by jointly learning to align and
	translate.
	\newblock {\em arXiv preprint arXiv:1409.0473}, 2014.
	
	\bibitem{lin2017structured}
	Zhouhan Lin, Minwei Feng, Cicero Nogueira~dos Santos, Mo~Yu, Bing Xiang, Bowen
	Zhou, and Yoshua Bengio.
	\newblock A structured self-attentive sentence embedding.
	\newblock {\em arXiv preprint arXiv:1703.03130}, 2017.
	
	\bibitem{tan2018deep}
	Zhixing Tan, Mingxuan Wang, Jun Xie, Yidong Chen, and Xiaodong Shi.
	\newblock Deep semantic role labeling with self-attention.
	\newblock In {\em Association for the Advancement of Artificial Intelligence},
	2018.
	
	\bibitem{vaswani2017attention}
	Ashish Vaswani, Noam Shazeer, Niki Parmar, Jakob Uszkoreit, Llion Jones,
	Aidan~N Gomez, {\L}ukasz Kaiser, and Illia Polosukhin.
	\newblock Attention is all you need.
	\newblock In {\em Advances in Neural Information Processing Systems}, pages
	5998--6008, 2017.
	
	\bibitem{li2018zoom}
	Hongyang Li, Yu~Liu, Wanli Ouyang, and Xiaogang Wang.
	\newblock Zoom out-and-in network with map attention decision for region
	proposal and object detection.
	\newblock {\em International Journal of Computer Vision}, pages 1--14, 2018.
	
	\bibitem{zhu2019attention}
	Yousong Zhu, Chaoyang Zhao, Haiyun Guo, Jinqiao Wang, Xu~Zhao, and Hanqing Lu.
	\newblock Attention couplenet: Fully convolutional attention coupling network
	for object detection.
	\newblock {\em IEEE Transactions on Image Processing}, 28(1):113--126, 2019.
	
	\bibitem{pirinen2018deep}
	Aleksis Pirinen and Cristian Sminchisescu.
	\newblock Deep reinforcement learning of region proposal networks for object
	detection.
	\newblock In {\em IEEE Conference on Computer Vision and Pattern Recognition},
	pages 6945--6954, 2018.
	
	\bibitem{yu2015multi}
	Fisher Yu and Vladlen Koltun.
	\newblock Multi-scale context aggregation by dilated convolutions.
	\newblock {\em International Conference on Learning Representations}, 2016.
	
	\bibitem{dai2017deformable}
	Jifeng Dai, Haozhi Qi, Yuwen Xiong, Yi~Li, Guodong Zhang, Han Hu, and Yichen
	Wei.
	\newblock Deformable convolutional networks.
	\newblock In {\em IEEE International Conference on Computer Vision}, pages
	764--773, 2017.
	
	\bibitem{huang2017densely}
	Gao Huang, Zhuang Liu, Laurens Van Der~Maaten, and Kilian~Q Weinberger.
	\newblock Densely connected convolutional networks.
	\newblock In {\em IEEE Conference on Computer Vision and Pattern Recognition},
	pages 4700--4708, 2017.
	
	\bibitem{fu2018dual}
	Jun Fu, Jing Liu, Haijie Tian, Zhiwei Fang, and Hanqing Lu.
	\newblock Dual attention network for scene segmentation.
	\newblock {\em arXiv preprint arXiv:1809.02983}, 2018.
	
	\bibitem{lin2014microsoft}
	Tsung-Yi Lin, Michael Maire, Serge Belongie, James Hays, Pietro Perona, Deva
	Ramanan, Piotr Doll{\'a}r, and C~Lawrence Zitnick.
	\newblock Microsoft coco: Common objects in context.
	\newblock In {\em European Conference on Computer Vision}, pages 740--755.
	Springer, 2014.
	
	\bibitem{girshick2018detectron}
	Ross Girshick, Ilija Radosavovic, Georgia Gkioxari, Piotr Doll{\'a}r, and
	Kaiming He.
	\newblock Detectron, 2018.
	
\end{thebibliography}
\end{document}